\documentclass[10pt,twocolumn,letterpaper]{article}

\usepackage[pagenumbers]{cvpr} % To force page numbers, e.g. for an arXiv version

\usepackage{graphicx}
\usepackage{amsmath}
\usepackage{amssymb}
\usepackage{booktabs}
\usepackage[utf8]{inputenc} % allow utf-8 input
\usepackage[T1]{fontenc}    % use 8-bit T1 fonts
\usepackage{url}            % simple URL typesetting
\usepackage{booktabs}       % professional-quality tables
\usepackage{amsfonts}       % blackboard math symbols
\usepackage{nicefrac}       % compact symbols for 1/2, etc.
\usepackage{microtype}      % microtypography
\usepackage{xcolor}         % colors

\usepackage{array}
\usepackage{booktabs} % For formal tables
\usepackage{multirow}  
\usepackage{mathtools}
\usepackage{enumerate}
\usepackage{graphicx}
\usepackage{colortbl}
\usepackage{amsmath}
\usepackage{amssymb}
\usepackage{bbm}
\usepackage{wrapfig}
\usepackage{caption}
\usepackage[accsupp]{axessibility} 
\usepackage{xspace}

\newcommand{\indicator}{\ensuremath{\mathbbm{1}}}

\newcommand{\denselist}{\itemsep 0pt\parsep=0pt\partopsep 0pt\vspace{-\topsep}}

\newenvironment{packed_itemize}
{\begin{itemize}
    \vspace{-\topsep}
    \setlength{\itemsep}{1pt}
    \setlength{\parskip}{0pt}
    \setlength{\parsep}{0pt}
}{\end{itemize}}

\newcommand{\nolistbottomspace}{\vspace{-\topsep}}

\usepackage[pagebackref,breaklinks,colorlinks]{hyperref}

\usepackage[capitalize]{cleveref}
\crefname{section}{Sec.}{Secs.}
\Crefname{section}{Section}{Sections}
\Crefname{table}{Table}{Tables}
\crefname{table}{Tab.}{Tabs.}

\def\cvprPaperID{7789} % *** Enter the CVPR Paper ID here
\def\confName{CVPR}
\def\confYear{2022}

\begin{document}

\title{The Neurally-Guided Shape Parser: \\
Grammar-based Labeling of 3D Shape Regions with Approximate Inference
}

\author{R. Kenny Jones\\
Brown University\\
\and
Aalia Habib\\
Brown University\\
\and
Rana Hanocka\\
University of Chicago\\
\and
Daniel Ritchie\\
Brown University\\
}
\maketitle

\begin{abstract}
We propose the Neurally-Guided Shape Parser (NGSP), a method that learns how to assign fine-grained semantic labels to regions of a 3D shape.
NGSP solves this problem via MAP inference, modeling the posterior probability of a label assignment conditioned on an input shape with a learned likelihood function.
To make this search tractable, NGSP employs a neural guide network that learns to approximate the posterior.
NGSP finds high-probability label assignments by first sampling proposals with the guide network and then evaluating each proposal under the full likelihood.
We evaluate NGSP on the task of fine-grained semantic segmentation of manufactured 3D shapes from PartNet, where shapes have been decomposed into regions that correspond to part instance over-segmentations. 
We find that NGSP delivers significant performance improvements over comparison methods that (i) use regions to group per-point predictions, (ii) use regions as a self-supervisory signal or (iii) assign labels to regions under alternative formulations.
Further, we show that NGSP maintains strong performance even with limited labeled data or noisy input shape regions.
Finally, we demonstrate that NGSP can be directly applied to CAD shapes found in online repositories and validate its effectiveness with a perceptual study. 
\end{abstract}

\section{Introduction}
\label{sec:intro}
 
The ability to semantically segment 3D shapes is important for numerous applications in vision, graphics, and robotics: reverse-engineering the part structure of an object to support editing and manipulation; producing training data for structure-aware generative shape models~\cite{gao2019sdmnet,StructureNet,jones2020shapeAssembly}; helping autonomous agents understand how to interact with objects in their environment~\cite{abbatematteo2019learning}; and more.
These applications often demand that the parts detected be fine-scale (e.g. wheels of an office chair) and hierarchically-organized (e.g. a cabinet door decomposes into a handle, door, and frame).
Producing such segmentations 
has proved to be a challenging task,
as it is expensive to gather large amounts of data at this granularity; PartNet~\cite{PartNet} is the only existing large-scale dataset of this type.

Recent work on 3D shape semantic segmentation has mainly focused on end-to-end approaches that operate on shape \textit{atoms} (e.g. mesh faces, point cloud points, occupancy grid voxels), i.e. the lowest-level geometric entity in the input representation~\cite{qi2017pointnet, qi2017pointnet++, MeshCNN, 3DShapeNets}. 
While these methods achieve impressive performance on many tasks, they do not often transfer well to domains with fine-grained labels or when access to labeled data is limited.
We postulate that one reason for this phenomenon is that attempting to label shape \textit{atoms} directly results in a massive search space, allowing learning-based methods to overfit unless the ratio of labeled shape instances to the label set complexity is high.

One way to address this issue is to design systems that make use of shape \textit{regions}.
When the number of shape \textit{regions} becomes significantly smaller than the number of shape \textit{atoms}, the label assignment problem becomes easier.
Such a framing may allow methods to learn fine-grained semantic segmentation when access to labeled data is limited. 
When shape \textit{regions} are provided, they can be used in various ways: (i) as a post-process aggregation on top of shape \textit{atom} predictions, (ii) to formulate auxiliary self-supervised objectives, or (iii) as the object to be labeled.
Methods that operate within this last paradigm can more directly reason about relationships \textit{between} regions, which can help improve fine-grained segmentation performance by better considering the context of a region within the entire shape.

The problem of decomposing a shape into regions useful for semantic segmentation is application-dependent. 
For CAD shapes and scenes found in online repositories, this type of region decomposition is often produced as a by-product of the modeling process, e.g. each part instance will be made out of one or more connected mesh components \cite{HierarchicalSegLabelOnline, SceneGrammar, LearningToGroupAndLabel}. 
Discovering region decompositions for shapes that do not already provide them is a well-studied problem within computer vision and graphics. There has been considerable recent effort on unsupervised techniques that approximate 3D shapes with primitives~\cite{tulsiani2017learning,AdaptiveHierarchicalCuboidAbstraction,deng2020cvxnet,NeuralStarDomain,NeuralParts}, and there is a long history of research on shape segmentation through purely geometric analysis~\cite{wang2011symmetry, asafi2013weak, kaick2014shape}.
There is even reason to believe that region decomposition solutions can generalize across shape categories, i.e. the way that shapes (especially manufactured objects) decompose into parts is largely category-independent~\cite{AdaCoSeg,Han20GenStruct}.

In this paper, we propose the Neurally-Guided Shape Parser (NGSP), a method that learns to assign fine-grained labels from a semantic grammar to regions of a 3D shape. 
Our approach is based on maximum \emph{a posteriori} (MAP) inference in a model of the probability that a label assignment to the shape's regions is correct.
Our likelihood consists of a mixture of modules that each operate on some regions of the shape. One set of modules evaluates the validity of the implied geometry and spatial layout for each label in the semantic grammar. 
Another module evaluates groups of regions formed by the label assignment.
As this combinatorial search problem is too complex to solve with exhaustive enumeration, we employ a neural guide network to approximate the posterior. 
The guide network reasons locally, predicting the label probability for each region independently.
Using the per-region probabilities produced by the guide network, NGSP importance samples a set of proposed label assignments.
To choose the best proposal out of this set, each label assignment is evaluated under the full likelihood, and the sample with highest posterior probability is chosen.

We compare NGSP against methods that use shape regions as a post-process, a self-supervisory signal, or assign labels to regions with different search strategies and likelihood formulations.
We evaluate each method on the task of fine-grained semantic segmentation of manufactured 3D shapes from PartNet, where each method has access to regions from the annotated part instance over-segmentations (e.g. each semantic part instance may consist of multiple regions).
NGSP achieves the best semantic segmentation performance, even in paradigms where access to labeled data is limited or when the input shape regions are noisy.
To validate our design decisions, we run an ablation study measuring the effect of each likelihood term and the neural guide network.
Finally, we show that NGSP can find good semantic segmentations on `in the wild' CAD shapes found from online repositories, and evaluate its performance with a forced choice perceptual study against comparison methods.
Code for our method and experiments can be found at found at https://github.com/rkjones4/NGSP .

In summary, our contributions are:
\begin{enumerate}[(i)]
\denselist
    \item We present the Neurally-Guided Shape Parser (NGSP), a method that learns how to assign labels from a semantic grammar to regions of a 3D shape. NGSP performs approximate MAP inference, using a guide network to find high-probability label assignments under a learned posterior probability of a label assignment conditioned on an input shape.
    \item We demonstrate that NGSP finds better fine-grained semantic segmentations for manufactured shapes compared with methods that use shape regions in alternative learning paradigms.
\denselist
\end{enumerate}

 \begin{figure*}[t!]
   \centering
   \includegraphics[width=\linewidth]{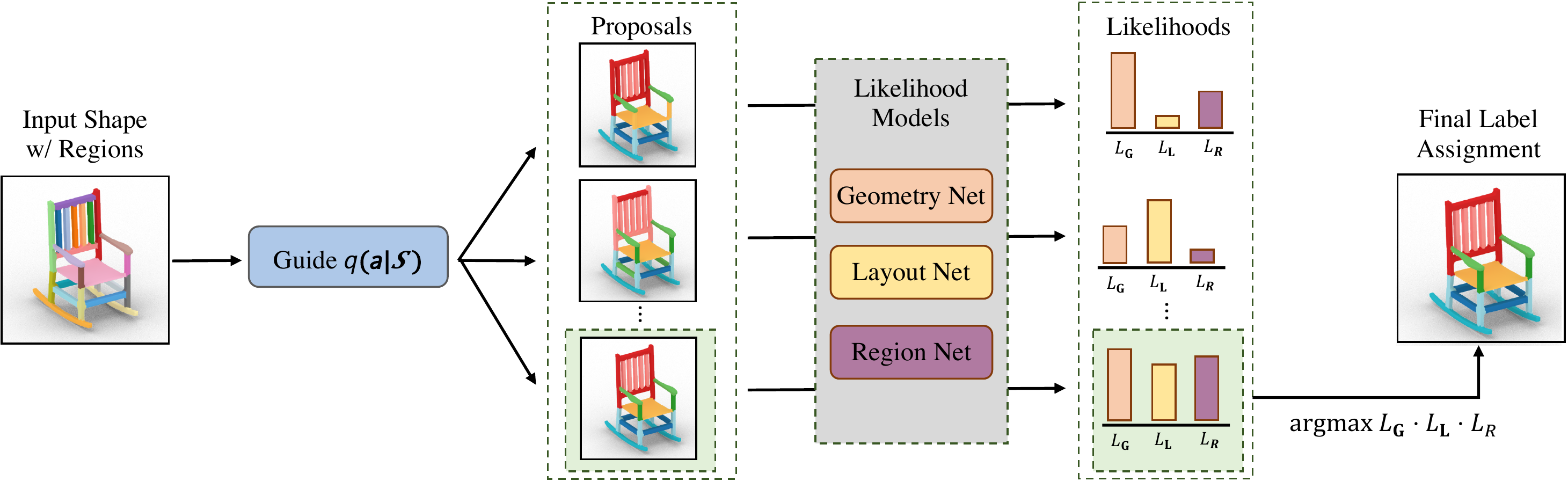}
\caption{The Neurally-Guided Shape Parser (NGSP) learns to assign fine-grained semantic labels (rightmost) to shape regions (leftmost). 
A guide network generates a set of proposed label assignments.
The label assignments are sent through likelihood modules that evaluate the global coherence of each proposal. 
These terms are combined into a posterior probability which determines the final label assignment.
}
 \label{fig:overview}
 \end{figure*}

\section{Related Work}
\label{sec:relwork}

{\bf Semantic Segmentation with 3D Shape Atoms}
Most learning-based methods for 3D shape semantic segmentation have used shape atoms (points, faces, edges, voxels) as their fundamental unit to label.
This practice dates back to pre-deep-learning work using conditional random fields on mesh faces~\cite{Kalogerakis:2010:labelMeshes} and extends to present-day, neural network methods including PointNet~\cite{qi2017pointnet}, PointNet++~\cite{qi2017pointnet++}, MeshCNN~\cite{MeshCNN}, and DGCNN~\cite{dgcnn}.
Some methods have been designed for settings where labeled data is limited, either in terms of the number of labels provided for each shape \cite{liu2021one, XuLee_CVPR20} or the number of shapes that contain any labels at all \cite{chen2019bae_net, selfsupacd, DBLP:conf/3dim/SharmaKM19}. 
While approaches within this paradigm achieve state-of-the-art performance for coarse, non-hierarchical segmentation, we show experimentally that they do not work as well in hierarchical, fine-grained settings where more inter-part relational reasoning is helpful.

{\bf Region-based Semantic Segmentation of Images and Scenes}
Our approach of decomposing a 3D shape into regions is conceptually similar to decomposing a 2D image into superpixels; there exist some prior work leveraging superpixels to improve image semantic segmentation.
Some of these methods use superpixels or other larger image regions to increase the computational efficiency of semantic segmentation~\cite{park2017superpixel} or to produce segmentation masks with crisper edges~\cite{xu2020refining,deformablegrid}.
A few of these methods, like ours, focus on achieving high accuracy with less training data~\cite{MultiLevelSuperpixelSeg,SuperpixelPooling,ClusteringSuperpixels}.

Similar ideas have also been proposed for segmenting 3D scenes. For large-scale scenes, points have been grouped into super-points to make learning approaches computationally tractable \cite{8590720,landrieu2018large}. 
Some 3D scene segmentation approaches explicitly compute labels per shape region.
One approach over-segments an indoor scene point cloud then uses a recursive denoising autoencoder to infer a hierarchical organization of those segments~\cite{HierarchyDenoisingRecursive}.
Another converts over-segmented indoor scenes into consistent hierarchies via dynamic-programming-based, bottom-up grammar parsing~\cite{SceneGrammar}.
The latter approach is similar to ours in that it also learns likelihoods from data; however, the scenes considered are more simplistic and easier to decompose into manageable sub-sections than the shapes we consider. In general, while scenes can be represented with point clouds, they have different characteristics from 3D shapes: scenes contain much fewer regular substructures and are more sparsely populated.

{\bf 3D Shape Semantic Hierarchies}
There is a long tradition of organizing 3D shapes and scenes into hierarchies.
Such hierarchies can be based on spatial locality or other metrics relating to convenience of editing and rendering, as in classical computer graphics.
One can also arrange part-based shapes into binary hierarchies based on connectivity and symmetry relationships between their parts~\cite{SymmetryHierarchy}; such a hierarchy can be a useful organization of shape data for training structure-aware generative models~\cite{GRASS}.
A generalization of this approach is to consider n-ary hierarchies; this is the data representation adopted by PartNet~\cite{PartNet}, which supports more sophisticated structure-aware generative shape models~\cite{StructureNet,jones2020shapeAssembly}.
Our method for semantic segmentation is designed with these kinds of hierarchies in mind and can help produce training data for such generative models.

{\bf Semantic Segmentation with 3D Shape Regions}
There has been some prior work that learns to assign semantic labels to 3D shape regions. 
One approach first learns how to group over-segmented shape regions from stock 3D models into part hypotheses, and then finds an optimal label assignment to each part hypothesis through a CRF formulation \cite{LearningToGroupAndLabel}. 
However, this method is not designed for hierarchical grammars, as it is unable to separate semantic parts that share similar bounding boxes, which is necessary for the fine-grained segmentations we desire (e.g. distinguishing a seat frame from a seat surface). 
Another approach proposes an MRF formulation where unary potentials capture per-region label probabilities and paired potentials encourage a smoothness term in relation to the grammar hierarchy \cite{HierarchicalSegLabelOnline}. 
We will show experimentally that NGSP outperforms this formulation on the task of fine-grained semantic segmentation.

Relatedly, some approaches have made use of a shape region decomposition to formulate self-supervised learning objectives.
One such method trains a PointNet++ to perform semantic segmentation, but also enforces a contrastive loss on per-point embeddings, encouraging points from the same shape region to share similar embeddings \cite{selfsupacd}. 
This technique achieves impressive performance on few-shot coarse segmentation tasks when a large collection of unsupervised shapes augments the labeled data set. 
We compare NGSP against this approach, and find that NGSP makes better use of shape regions for fine-grained segmentations, even with limited labeled data.

\newcommand{\shape}{\ensuremath{\mathcal{S}}}
\newcommand{\region}{\ensuremath{R}}
\newcommand{\regionGroup}{\ensuremath{\mathcal{R}}}

\newcommand{\grammar}{\ensuremath{\mathcal{G}}}
\newcommand{\slabels}{\ensuremath{L}}
\newcommand{\terminals}{\ensuremath{\slabels_\mathbf{T}}}
\newcommand{\nonterminals}{\ensuremath{\slabels_\mathbf{V}}}
\newcommand{\slabel}{\ensuremath{l}}
\newcommand{\terminal}{\ensuremath{\slabel_\mathbf{T}}}
\newcommand{\nonterminal}{\ensuremath{\slabel_\mathbf{V}}}
\newcommand{\axiom}{\ensuremath{\omega}}
\newcommand{\productions}{\ensuremath{P}}
\newcommand{\assignments}{\ensuremath{\mathbf{A}}}
\newcommand{\assignment}{\ensuremath{\mathbf{a}}}
\newcommand{\likelihood}{\ensuremath{\mathcal{L}}}

\newcommand{\likeG}{\ensuremath{\mathcal{L}_\mathbf{G}}}
\newcommand{\likeL}{\ensuremath{\mathcal{L}_\mathbf{L}}}
\newcommand{\likeR}{\ensuremath{\mathcal{L}_\mathbf{R}}}
\newcommand{\likeQ}{\ensuremath{\mathcal{L}_\mathbf{Q}}}

\section{Method}
\label{sec:method}

The input to our method is a shape $\shape$ which has been decomposed into a set of regions $\region$, i.e. $\shape = \{ \region_i \}$.
Our method also receives as input a label grammar $\grammar = (\slabels, \axiom, \productions)$, where $\slabels$ is a set of possible semantic labels, $\axiom$ is the root label (the \emph{axiom} of the grammar), and $\productions \subset \slabels \times \slabels^*$ is the set of production rules for the grammar (specifying which labels can be the children of other labels).
The label set $\slabels$ can be divided into \emph{terminal} labels $\terminals$ (those with no children) and \emph{non-terminal} labels $\nonterminals$, such that $\slabels = \terminals \cup \nonterminals$.
We assume that there exists a unique path from the root to each terminal label $\terminal \in \terminals$, i.e. every label has at most one parent.
This is a reasonable assumption for shape labeling; all PartNet~\cite{PartNet} label grammars have this property.

  \begin{figure*}[t!]
   \centering
   \includegraphics[width=\linewidth]{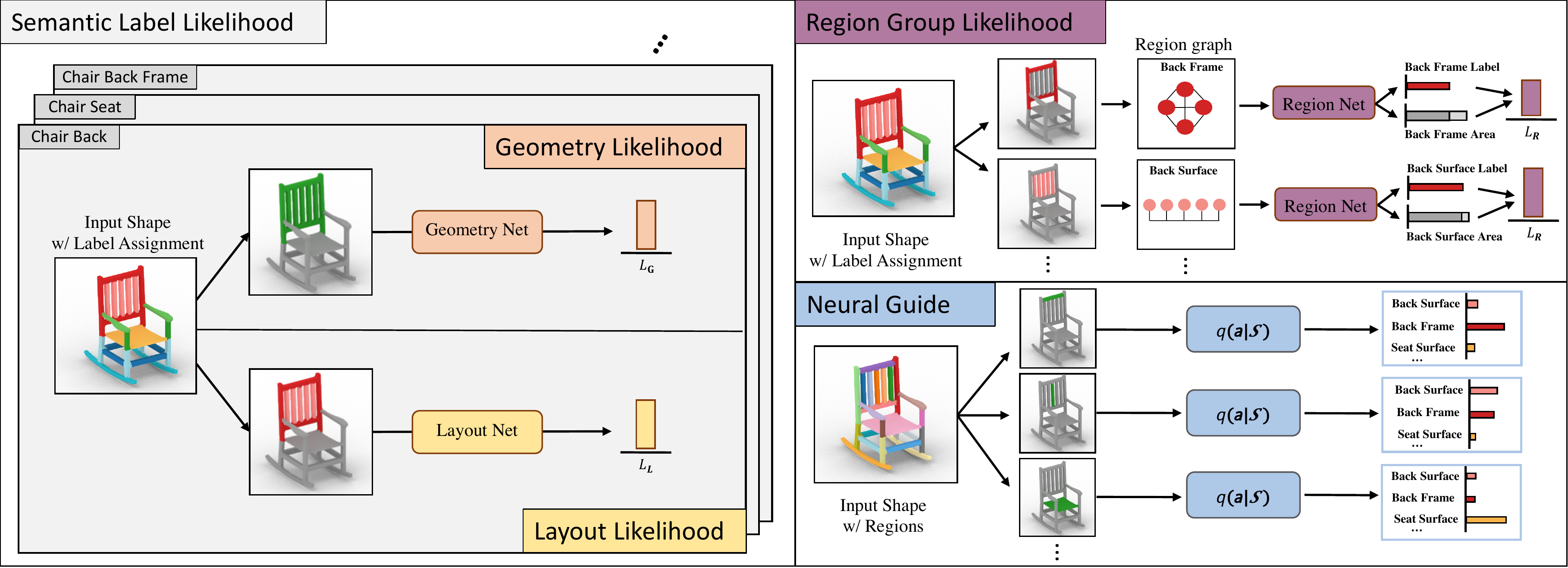}
\caption{ Design of NGSP's modules. The geometry and layout likelihoods consume a (shape, label assignment) pair, and are computed for each semantic label in the grammar \emph{(left)}. Each geometry network sees which regions of the input shape have been assigned to its label (e.g. chair back). Each layout network sees which regions of the input shape have been assigned to its child labels (e.g. chair back surface and chair back frame). The region group likelihood term also takes a (shape, label assignment) pair as input \emph{(top-right)}. For each group of regions implied by the label assignment, it creates a fully-connected graph, where nodes correspond to shape regions in the group. The neural guide network operates over individual shape regions, predicting the label for each region independently \emph{(bottom-right)}}.
 \label{fig:architects}
 \vspace{-5.0mm}

 \end{figure*}

Given these inputs, our goal is to find the maximum \emph{a posteriori} (MAP) label assignment $\assignments = \{ \assignment_i \}$, where $\assignment_i = \assignments(\region_i)$ is the label assigned to region $\region_i \in \shape$.
We assume a uniform prior distribution over labels and model the posterior $p(\shape | \assignments)$ with a data-driven likelihood function:
\vspace{-2.0mm}
\begin{equation}
\label{eq:lmap}
\likelihood(\shape, \assignments) = \likeG(\shape, \assignments) \cdot \likeL(\shape, \assignments) \cdot \likeR(\shape, \assignments)
\vspace{-2.0mm}
\end{equation}
$\likeG$ and $\likeL$ reason about properties of the semantic labels of $\grammar$, while $\likeR$ reasons about properties of groups of regions implied by a given assignment. 

As the search space of label assignments to shape regions is large, especially with fine-grained label sets, we guide our search with a network that learns to locally approximated the posterior: $q(\assignment | \shape)$.
Figure~\ref{fig:overview} outlines our approach.
Using this guide network, we importance sample a set of complete label assignments, which we call proposals. These proposals are then evaluated under Equation~\ref{eq:lmap}, and the proposal that returns the highest likelihood is chosen as the final output label assignment.

In the remainder of this section, we describe the different components of this pipeline in more detail: the semantic label likelihood terms (Section~\ref{sec:sem_lik}), the region group likelihood term (Section~\ref{sec:reg_lik}), and details of our neurally-guided search procedure (Section~\ref{sec:prior}).

\subsection{Semantic Label Likelihood Terms}
\label{sec:sem_lik}

For each label in the grammar, the semantic label likelihood terms reason about different properties of shape regions that were assigned to that label.
Specifically, for each $l \in \grammar$, we learn to identify geometric properties of $l$ with a geometry likelihood $\likeG$ and semantic layout properties of $l$ with a layout likelihood $\likeL$.
$\likeG$ aims to capture information about the typical geometric properties of regions assigned to a given label (e.g. chair seats usually have a flat top surface); $\likeL$ aims to capture typical spatial relationships between a label's children (e.g. within a chair base, a rocker is usually positioned underneath the legs). Both of these likelihoods are modeled with the same structure:
\begin{align}
    &\likeG(\shape, \assignments) = \Big( \prod_{\slabel \in \slabels} p_\mathbf{G}(\shape_{\slabel : \assignments} | \slabel) \Big)^{1/(\sum_{\slabel \in \slabels} \indicator[ \slabel \in \assignments ]) } \nonumber \\
    &\likeL(\shape, \assignments) = \Big( \prod_{\slabel \in \slabels} p_\mathbf{L}(\shape_{\slabel : \assignments} | \slabel) \Big)^{1/(\sum_{\slabel \in \slabels} \indicator[ \slabel \in \assignments ]) } \nonumber\\
    &\shape_{\slabel : \assignments} = \{ \region \in \shape \text{\ s.t.\ } \assignments(\region) = \slabel \} \nonumber
\end{align}

where $\shape_{\slabel : \assignments}$ is the subset of regions in the shape $\shape$ which are assigned to label $\slabel$ in assignment $\assignments$.
The exponents normalize these probabilities by the number of labels which occur in the label assignment $\assignments$, e.g. the number of non-unity product terms.

We model the geometry network $p_\mathbf{G}(\shape_{\slabel : \assignments} | \slabel)$ and layout network $p_\mathbf{L}(\shape_{\slabel : \assignments} | \slabel)$ with PointNet++ architectures, where each input point cloud contains surface samples from  $\shape_{\slabel : \assignments}$ (Figure \ref{fig:architects}, left).
Conditioning on $\slabel$ is implemented by training separate $p_\mathbf{G}$ and $p_\mathbf{L}$ networks for each label $\slabel \in \slabels$. 

Each network is trained in a binary classification paradigm, tasked with assessing whether the regions in $\shape_{\slabel : \assignments}$ are a valid instance of a semantic part with label $\slabel$.
Positive examples are sourced from the training dataset: the networks for label $\slabel$ receive one positive example of $\shape_{\slabel : \assignments}$ from each shape where $\slabel$ appears. 
Negative examples come from synthetically-generated corruptions of each positive example (i.e. changing region labels).
To encourage the geometry and layout networks to focus on the properties after which they are named, we introduce the following inductive biases (details in the supplemental):

{\bf Geometry Network:}
The geometry network should learn to reason about whether the shape of the union of regions in $\shape_{\slabel : \assignments}$ is consistent with the label $\slabel$.
Thus, each negative example is derived by adding or removing regions from a positive example.

{\bf Layout Network:}
The layout network should focus on whether the relationships between the child labels of regions assigned to $\slabel$ are consistent with that label.
To enable this reasoning, the network receives the child label as an additional one-hot attribute concatenated to every point.
Each negative example is derived by modifying the child label assignment of at least one region from a positive example. 

\subsection{Region Group Likelihood Term}
\label{sec:reg_lik}

The region group likelihood term reasons about properties of region groups implicitly formed when a labeling is assigned to an input shape.
Specifically, it models the probability that ($\shape$, $\assignments$) pairs are valid with respect to region groups $\regionGroup$ of $\shape$ formed under $\assignments$. 
For each $\terminal \in \assignments$, the region group $\regionGroup_l$ is defined to be \{$\region_{i} \in \shape \mid \assignment_{i} = \terminal \}$.

$\likeR$ reasons over two properties of each $\regionGroup_l$: if $l$ is the best label for $\regionGroup_l$ and what percentage of area within $\regionGroup_l$ belongs to $l$.
We model these properties with a region network $p_\mathbf{R}$. It consumes a region group $\regionGroup_l$,
and predicts the probability that $\regionGroup_l$ has $l$ as its majority label, $p^\text{label}_{\mathbf{R}}$, and the percentage of the area within $\regionGroup$ that has $l$ as its true label, $p^\text{area}_{\mathbf{R}}$. 
These predictions are then combined and normalized across all region groupings:
\vspace{-2.5mm}
\begin{equation*}
\likeR(\shape, \assignments) = \Big( \prod_{\slabel \in \slabels} p^\text{label}_{\mathbf{R}}(\regionGroup_l | \slabel) \cdot  p^\text{area}_{\mathbf{R}}(\regionGroup_l | \slabel) \Big)^{1/|\regionGroup|}
\vspace{-2.5mm}
\end{equation*}
We model $p_\mathbf{R}$ with a region-based graph convolutional network (Figure \ref{fig:architects}, top-right). 
We convert each $\regionGroup_l$ into a fully-connected graph where the nodes correspond to the regions of $\regionGroup_l$. 
We initialize node and edge features with embeddings predicted by a pretrained point cloud auto-encoder; details are provided in the supplemental.
$p_\mathbf{R}$ performs 4 rounds of gated graph convolution, then creates a single latent representation for the entire graph with a max-pooling layer \cite{dwivedi2020benchmarkgnns, bresson2017residual}.
$p^\text{label}_{\mathbf{R}}$ is modeled with a linear layer that predicts a probability distribution over the terminal label set.
$p^\text{area}_{\mathbf{R}}$ is conditioned on $l$ and modeled with a linear layer that predicts a scalar value in $[0,1]$, where 0 implies none of the area within $\regionGroup_l$ belongs to $l$ and 1 implies all of the area within $\regionGroup_l$ belongs to $l$.

\subsection{Neurally-Guided Search}
\label{sec:prior}

While the search space over regions is much smaller than the search space over atoms, it is still computationally infeasible to exhaustively evaluate $\likelihood$ on all possible label assignments to regions. To guide our search procedure towards good areas of the search space, we learn a guide network $q(\assignment | \shape)$ to locally approximate the posterior.

We model $q(\assignment | \shape)$ with a neural network trained to predict the probability of each possible label assignment $\assignment_i$ for each region $\region_i$ of the shape $\shape$.
$q(\assignment | \shape)$ uses a PointNet++ architecture \cite{qi2017pointnet++}, where the input point cloud contains samples from the entire shape, but each point has an extra one-hot dimension indicating whether it belongs to the region of interest (Figure \ref{fig:architects}, bottom-right). We train $q(\assignment | \shape)$ in a classification paradigm, where each shape $\shape$ in the dataset produces $|\shape|$ training examples (one for each region), and the classification target for each example is the ground truth semantic label of that region. We can then calculate the approximate posterior guide probability, $\likeQ$, of a ($\shape$,$\assignments$) pair with the following equation:

\vspace{-3.0mm}
\begin{equation*}
    \likeQ(\shape, \assignments) = \prod_{i=1}^{|\shape|} q(\assignment_i) 
    \vspace{-2.5mm}
\end{equation*}

At inference time, our goal is to find high likelihood label assignments $\assignments$ for a given shape $\shape$. 
To achieve this, our procedure creates a set of proposed label assignments by using $q(\assignment | \shape)$ to importance sample the top $k$ label assignments to $\shape$ under $\likeQ$. 
We then evaluate each proposed assignment under $\likelihood$ and select the label assignment within this set which maximizes Equation~\ref{eq:lmap}.

\section{Experiments}
\label{sec:results}

In this section, we evaluate NGSP's ability to assign semantic labels to regions of 3D shapes. 
Our experiments use CAD manufactured objects from the PartNet dataset \cite{PartNet} (Section \ref{sec:res_data}). 
We describe the details of our training procedure in Section \ref{sec:res_regime}. 
In Section \ref{sec:res_sem_seg}, we compare NGSP against region-aware comparison methods on the task of semantic segmentation under varying amounts of labeled training data.
We provide an ablation study on the components of NGSP in Section \ref{sec:res_ablation}. 
We examine how NGSP is affected when input shape regions are artificially corrupted (Section \ref{sec:res_corrupt}) or are produced by an ACD method (Section \ref{sec:acd_exp}).
Finally, in Section \ref{sec:perceptual_study} we run NGSP on `in the wild' CAD shapes, and compare its predicted segmentations against alternative methods with a forced choice perceptual study.

\subsection{Data}
\label{sec:res_data}

We consider six categories of manufactured shapes from PartNet~\cite{PartNet}: chairs, lamps, tables, storage furniture, vases, and knives. 
We use PartNet's hierarchical labelings as our ground truth: on average, each label grammar contains 34 total labels and 21 leaf labels.
The dataset for each category contains between 300 and 1200 shapes, split between train, validation and test sets.
We over-segment each shape using the mesh components for each part instance in PartNet (a part instance may consist of multiple components).
For training and inference, we convert each mesh into point clouds with a surface sampling.
Full details are provided in the supplemental material.

\subsection{Training Details}
\label{sec:res_regime}

\begin{table}[t!]
    \setlength{\tabcolsep}{2pt}
    \centering    
    \footnotesize
    \begin{tabular}{@{}cl c   cccccc@{}}
        \toprule
        \textbf{\# Train} & \textbf{Method} &  \textbf{Mean} & \textbf{Chair} & \textbf{Lamp} & \textbf{Table} & \textbf{Vase} & \textbf{Knife}  &\textbf{Storage}  \\
        \midrule
        \multirow{5}{*}{\emph{10}}
        & PartNet (R) & 18.1  & 25.3& 10.2 & 3.2 & 12.6 & 33.2 & 24.2  \\
        & BAE-NET (R) & 20.7 & 23.3  & 10.7 & 11.0 & 35.7 & 22.2 & 21.8  \\
        & LEL (R) & 20.1 & 31.1  & 14.3 & 8.6 & 12.6 & 27.4 & 26.8  \\
        & LHSS & 24.3 & 24.7        & 16.7 & 13.0 & 33.3 & \textbf{34.1} & 23.9\\
        & NGSP & \textbf{33.6} &\textbf{ 36.6 }       & \textbf{24.7 }& \textbf{16.3} & \textbf{58.8} & 29.3 &  \textbf{35.9} \\
        \midrule
        \multirow{5}{*}{\emph{40}}
        & PartNet (R) & 31.6  & 39.4& 24.5 & 19.1 & 44.9 & 25.5 & 36.0  \\
        & BAE-NET (R) & 26.5 & 30.5  & 19.0 & 13.1 & 42.4 & 27.9 & 25.9 \\
        & LEL (R) & 38.6 & 45.4  & 26.4 & 26.1 & 48.0 & 45.3 & 40.3  \\
        & LHSS & 35.4 & 35.7        & 23.3 & 20.1 & 50.0 & 44.3 & 39.1  \\
        & NGSP & \textbf{50.9}& \textbf{53.6} & \textbf{42.8} & \textbf{30.4} & \textbf{76.2} & \textbf{49.7} & \textbf{52.9}  \\
        \midrule
        \multirow{5}{*}{\emph{400}}
        & PartNet (R) & 41.2 & 49.0 & 24.6 & 37.8 & 53.9 & 42.1 & 39.9  \\
        & BAE-NET (R) & 30.4 & 34.7  & 29.6 & 16.6 & 44.3 & 28.7 & 28.3  \\
        & LEL (R) & 41.9 & 48.0  & 38.0 & 38.2 & 46.4 & 41.2 & 39.4  \\
        & LHSS    &  36.3   & 43.7  & 29.0 & 31.2 & 45.0 & 33.1 & 36.0 \\
        & NGSP & \textbf{57.9} & \textbf{63.6}        & \textbf{44.6} & \textbf{45.3} & \textbf{84.6} & \textbf{55.9} & \textbf{53.2} \\
        \bottomrule
    \end{tabular}
    \caption{Fine-grained semantic segmentation results across different PartNet categories. The metric is mIoU (higher values are better). NGSP significantly outperforms other methods that make alternative use of shape regions. This trend remains consistent even in limited labeled data regimes (\# Train column).}
    \vspace{-3.0mm}
    \label{tab:seg_comp}
\end{table}

The layout, geometry, and region label networks are trained with binary cross entropy. The region area network is trained with L1 loss. The guide network is trained with focal cross entropy loss \cite{xu2014focal}.
We use the Adam optimizer \cite{Kingma2014AdamAM} with a learning rate of $10^{-3}$ for the guide network and $10^{-4}$ for all other networks. 
All networks perform early stopping using the validation set. 
Models were trained sequentially on a machine with a GeForce RTX 2080 Ti GPU with an Intel i9-9900K CPU, consuming up to 10GB of GPU memory and taking between 1-2 days to train for the categories with more semantic labels.
See the supplemental material for full details about network architectures.

\subsection{Fine-Grained Semantic Segmentation}
\label{sec:res_sem_seg}

We compare NGSP against alternative region labeling methods on the task of semantic segmentation.
All evaluations are performed on a held-out test set. 
Unless otherwise stated, the number of sampled proposals from the guide network, $k$, is set to 10000.
Following PartNet, we use mIoU as our evaluation metric: the intersection over union between predicted and ground-truth per-point labels, averaged over labels in the grammar.
  
We compare NGSP to the following methods. Methods appended by (R) make per-point predictions which are aggregated with an average operation into per-region predictions to form a full label assignment. 
\begin{packed_itemize}
    \item \textbf{PartNet (R)}: De-facto approach for fine-grained semantic segmentation that uses a PointNet++ to predict into the terminal label set \cite{PartNet}.
    \item \textbf{BAE-NET (R)}: Implicit field network that jointly learns to semantically segment and reconstruct shapes; designed for limited labeled data \cite{chen2019bae_net}.
    \item \textbf{LEL (R)}: PointNet++ back-bone where shape region decompositions formulate a self-supervised training objective augmenting the classification loss; designed for limited labeled data \cite{selfsupacd}.
    \item \textbf{LHSS}: Constructs an MRF where nodes correspond to shape regions. Finds low-cost label assignments over learned unary and grammar-based pairwise potentials with an alpha-expansion algorithm \cite{HierarchicalSegLabelOnline}.
\nolistbottomspace
\end{packed_itemize}
Each method is trained with access to the same labeled shape instances.
BAE-NET and LEL are additionally provided with up to 1000 shape instances per class that lack semantic label annotations but contain region decompositions.
Full details are provided in the supplemental.

{\bf Results:} Quantitative results of this experiment are shown in Table~\ref{tab:seg_comp}.
When labeled data is plentiful (400 max training shapes, bottom rows), NGSP outperforms the comparison methods by a significant margin. 
Looking at the mean result across categories, NGSP offers a 38\% improvement over the next best method (LEL). 
When access to labeled data is limited, NGSP also outperforms alternatives with a 31\% improvement when 10\% of the training data is used and a 38\% improvement when 2.5\% of the training data is used. 
In fact, NGSP's mean category performance with 10\% of the labeled data outperforms any comparison method that has access to all of the labeled data by almost 10 absolute percentage points. 
This result suggests that NGSP could be useful for semantic segmentation of 3D shapes from uncommon categories for which datasets of semantically annotated instances are not readily available.

We present some qualitative comparisons from the same experiment in Figure~\ref{fig:qual_seg_comp}, and provide additional examples in the supplemental.
NGSP is able to find label assignments that are more coherent, and better reflect the ground-truth labels, compared with the alternative methods. 
Methods that rely on regions to group per-atom predictions often produce segmentations that lack global consistency.
LHSS attempts to reason about global consistency with its pairwise potentials, but these encourage the output segmentation to become overly smooth, missing fine-grained part distinctions.

  \begin{figure*}[t!]
    \centering
    \setlength{\tabcolsep}{1pt}
    \begin{tabular}{ccccccc}
        Input Regions & PartNet (R) & BAE-NET (R) & LEL (R) & LHSS & NGSP & GT
        \\
        \includegraphics[{width=.14\linewidth}]{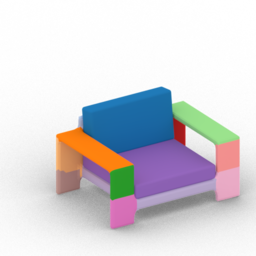} &
        \includegraphics[{width=.14\linewidth}]{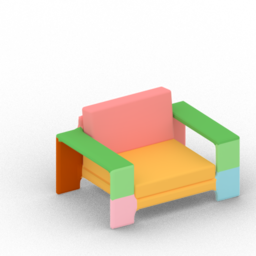} &
        \includegraphics[{width=.14\linewidth}]{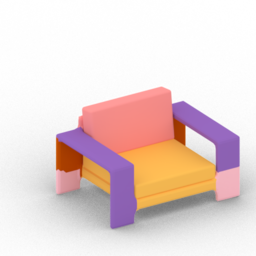} &
        \includegraphics[{width=.14\linewidth}]{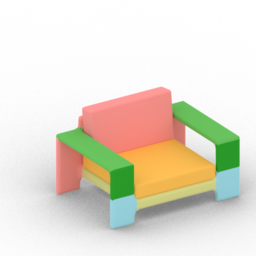} &
        \includegraphics[{width=.14\linewidth}]{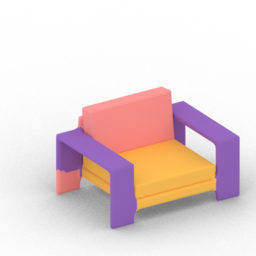} &
        \includegraphics[{width=.14\linewidth}]{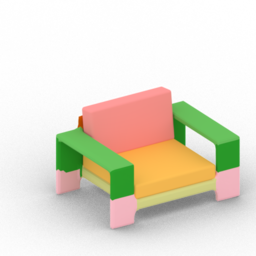} &
        \includegraphics[{width=.14\linewidth}]{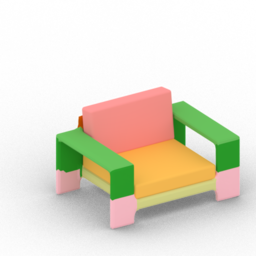} \\
        
        \includegraphics[{width=.14\linewidth}]{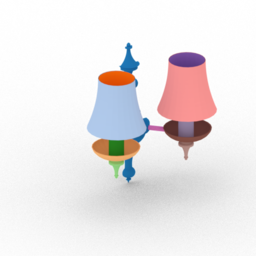} &
        \includegraphics[{width=.14\linewidth}]{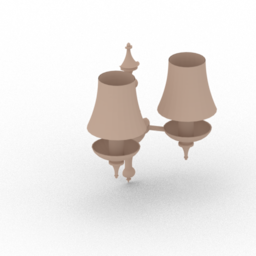} &
        \includegraphics[{width=.14\linewidth}]{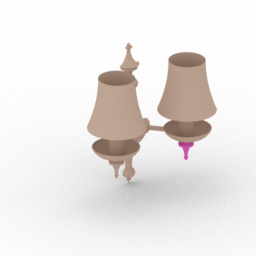} &
        \includegraphics[{width=.14\linewidth}]{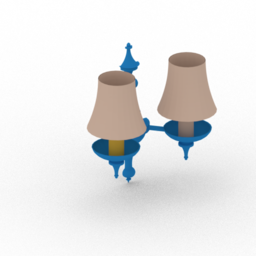} &
        \includegraphics[{width=.14\linewidth}]{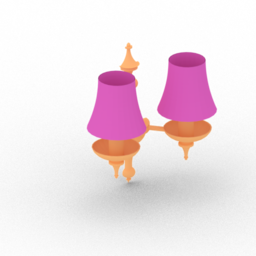} &
        \includegraphics[{width=.14\linewidth}]{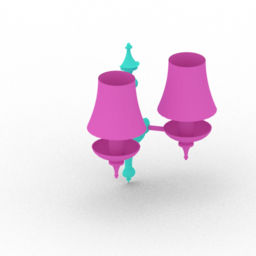} &
        \includegraphics[{width=.14\linewidth}]{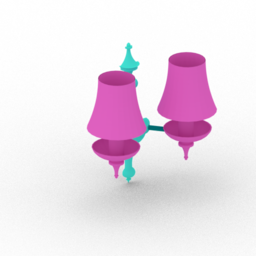} \\
        
        \includegraphics[{width=.14\linewidth}]{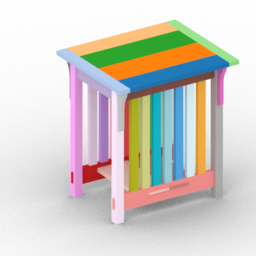} &
        \includegraphics[{width=.14\linewidth}]{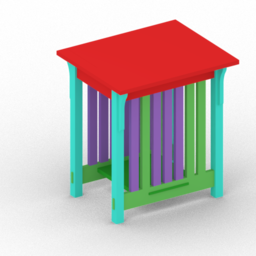} &
        \includegraphics[{width=.14\linewidth}]{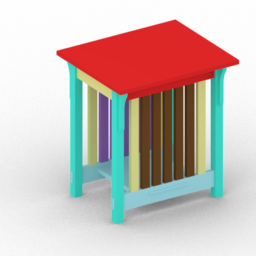} &
        \includegraphics[{width=.14\linewidth}]{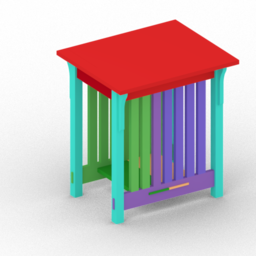} &
        \includegraphics[{width=.14\linewidth}]{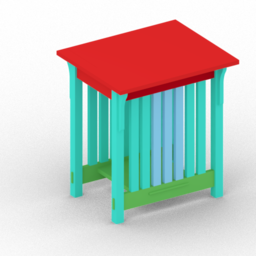} &
        \includegraphics[{width=.14\linewidth}]{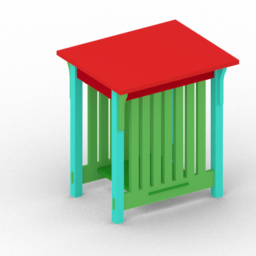} &
        \includegraphics[{width=.14\linewidth}]{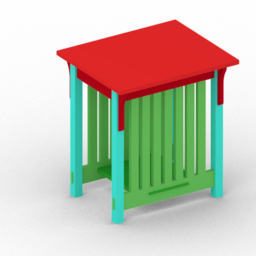} \\
    \end{tabular}
    \caption{ Qualitative comparison of fine-grained semantic segmentations. We show the input shape regions \emph{(left)}, the ground-truth label assignment \emph{(right)}, and the label assignments produced by different methods \emph{(middle)}. Each semantic label is represented by a unique color. NGSP predicts label assignments that best agree with the ground-truth. We present additional qualitative results in the supplemental. }
    \label{fig:qual_seg_comp}
\end{figure*}

\subsection{Ablation Study}
\label{sec:res_ablation}

\begin{table}[]
    \setlength{\tabcolsep}{5pt}
    \footnotesize
    \centering    
    \begin{tabular}{@{}l ccc@{}}
        \toprule
       \textbf{Model} & \textbf{10 Train} & \textbf{40 Train} & \textbf{400 Train} \\
        \midrule
        No $\likeG$ & 30.7 & 47.2 & 57.3 \\
        No $\likeL$ & 29.0 & 46.1 & 56.3 \\
        No $\likeR$ & 32.7 & 48.0 & 54.0 \\
        No $\likelihood$ & 29.3 & 43.0 & 51.6 \\
        No $q(\assignment | \shape)$ & 11.7 & 13.3 & 13.0  \\
        NGSP & \textbf{33.6 }& \textbf{50.9 }&  \textbf{57.9}  \\
        \bottomrule
    \end{tabular}
    \caption{Semantic segmentation performance of NGSP under different ablation conditions (metric is mIoU, averaged across categories). Each component of NGSP helps it find good label assignments.
    } 
    \vspace{-3.0mm}
    \label{tab:ablation}
\end{table}

To evaluate the design of NGSP, we conduct a series of ablations, where each formulation has one component of NGSP removed:
\begin{packed_itemize}
    \item \textbf{No $\likeG$}: Geometry likelihood is removed from $\likelihood$.
    \item \textbf{No $\likeL$}: Layout likelihood is removed from $\likelihood$.
    \item \textbf{No $\likeR$}: Region group likelihood is removed from $\likelihood$.
    \item \textbf{No $\likelihood$}: The best proposal under $\likeQ$ is chosen.
    \item \textbf{No $q(\assignment | \shape)$}: $\likelihood$ evaluates proposals from a uniform prior.
\end{packed_itemize}

We present results of this experiment in Table \ref{tab:ablation}.
As we show across multiple training set sizes, removing any component of $\likelihood$ (top 3 rows) leads to a worse mIoU. 
The ``No $q(\assignment | \shape)$'' row demonstrates the importance of the neural guide network: the search space is too large to effectively explore in a naive manner.
However, as seen in the ``No $\likelihood$'' row, the predictions of $q(\assignment | \shape)$ can be furthered improved by evaluating its proposals under a better estimate of the posterior.
As $q(\assignment | \shape)$ only evaluates regions locally, it is unable to benefit by reasoning about part-to-part relationships implied by the global label assignment in the same way as $\likelihood$.

\subsection{Sensitivity to Region Corruption}
\label{sec:res_corrupt}

\begin{table}[]
    \footnotesize
    \centering
        \begin{tabular}{@{}lccc@{}}
            \toprule
            \textbf{Method} & \textbf{1X Reg} & \textbf{2X Reg} & \textbf{4X Reg} \\
            \midrule
            PartNet (R) & 41.2 & 40.7 &  40.7\\
            BAE-NET(R) & 30.4 & 30.3 & 29.9\\
            LEL(R) & 41.9 & 41.7 & 41.3 \\
            LHSS & 36.3 & 35.9 & 35.4 \\
            NGSP  & \textbf{57.9}  & \textbf{49.0} & \textbf{45.3}\\
            \bottomrule
        \end{tabular}
        \caption{ We evaluate the semantic segmentation performance of different methods in regimes where shape regions have undergone artificial corruption (metric is mIoU, averaged across categories).   
        NGSP's performance declines gradually as the corruption increases, but in all cases remains better than alternative methods.}
        \vspace{-3.0mm}
        \label{tab:corrupt}
\end{table}

We analyze how sensitive NGSP is to corruptions of the part instance over-segmented regions of the input shapes.
For this analysis, we construct datasets of shapes whose regions have been artificially split into smaller sub-regions.
In the 2X (4X) paradigm, each region is split into 2 (4) regions; details of how these splits are produced are provided in the supplemental.
For each corruption paradigm, the neural guide network is retrained on training shapes whose regions have undergone similar corruption.
Results of this experiment are shown in Table~\ref{tab:corrupt}, where we track semantic segmentation performance against baselines which receive the same corrupted regions.
As the amount of region corruption increases, the performance of NGSP declines, but in every condition it continues to offer performance improvements over all comparison methods.

\subsection{Applications to Unstructured Data}
\label{sec:acd_exp}

As NGSP requires a region decomposition as input, it can’t be directly applied to some types of unstructured data without the help of auxiliary methods. 
While there are many methods that aim to convert unstructured shape data into a reasonable region decomposition, all existing methods have limitations, and this remains a hard, unsolved problem.
However, even though these region decompositions may contain errors, NGSP can still use them to improve semantic segmentation performance when access to labeled data is limited. 
We run an experiment comparing NGSP against alternative region labeling methods over unstructured input data, with regions created by the ACD method from \cite{selfsupacd}. We report the mean category mIoU that each method achieves with ACD produced regions while training over 10 training shapes (Table \ref{tab:acd_res}). In this paradigm, NGSP makes the best use of the ACD regions, but all methods perform worse compared with using the PartNet provided regions (Table \ref{tab:seg_comp}). 

\begin{table}[]
        \centering
        \footnotesize
        \begin{tabular}{@{}lc@{}}
            \toprule
            \textbf{Method} & \textbf{Mean mIoU} \\
            \midrule
            PartNet + NR & 0.155 \\
            PartNet + ACD & 0.161  \\
            BAE-NET + ACD  & 0.180 \\
            LEL + ACD  & 0.206  \\
            LHSS + ACD  &  0.202 \\
            NGSP + ACD   & \textbf{0.244} \\
            \bottomrule
        \end{tabular}
    %\vspace{-2.0mm}
    \caption{Semantic segmentation performance over unstructured input data with ACD generated regions and 10 labeled training shapes (NR is no regions).}
    \label{tab:acd_res}
\end{table}

\subsection{Applications to `in the wild' CAD Shapes}
\label{sec:perceptual_study}

As a byproduct of CAD modeling procedures, many `in the wild' 3D shapes come with part instance over-segmentations.
NGSP can segment such objects by treating each mesh connected component as a shape region.
To demonstrate this application, we compile a small dataset of 26 meshes from the chair category of ShapeNet, where each shape's connected components form a reasonable approximation to a part instance over-segmentation.
We run NGSP and two comparison methods (PartNet (R) and LHSS) on each shape and record each method's predicted label assignment.
As we lack ground-truth label annotations for these shapes, we evaluate NGSP with a two-alternative forced choice perceptual study.
Each participant was shown a sequence of examples, where each example visualized two ways that parts of a chair could be labeled, and was asked to select the part labeling that better matched the given shape. 
Further details provided in the supplemental.

\textbf{Results} We present the results of this perceptual study in Table \ref{tab:perceptual}. 
Participants had a strong preference for the part labelings generated by NGSP.
In comparisons against PartNet, NGSP was preferred 79.1\% on average, with a 95\% confidence interval lower-bound of 66.1\%.
In comparisons against LHSS, NGSP was preferred 79.6\% on average, with a 95\% confidence interval lower-bound of 68.1\%.

\begin{table}[]
    \small
    \centering
        \begin{tabular}{@{}lcc@{}}
            \toprule
            \textbf{NGSP vs.} & \textbf{Mean} & \textbf{95\% CI} \\
            \midrule
            PartNet (R) & 79.1 & [66.1, 92.1] \\
            LHSS & 79.6 & [68.1, 91.1] \\
            \bottomrule
        \end{tabular}
        \caption{Quantitative results of our perceptual study comparing semantic segmentations produced by different methods on `in the wild' CAD shapes. NGSP's label assignments were significantly preferred over those predicted by Partnet (R) or LHSS.}
        \label{tab:perceptual}
        \vspace{-4.0mm}
\end{table}

\section{Conclusion}
\label{sec:conc}

We presented the Neurally-Guided Shape Parser (NGSP), a method that performs semantic segmentation on region-decomposed 3D shapes.
NGSP assigns labels to shape regions via MAP inference in a learned model of the probability that a label assignment is correct conditioned on the shape's regions.
Search is made tractable through an approximate inference scheme, where the exploration of label assignments is constrained by a neural guide network.
We experimentally demonstrated that NGSP outperforms methods that (i) use regions to aggregate point predictions (ii) incorporate regions into self-supervised training objectives or (iii) assign labels to regions in alternative search-based formulations. 
We observed that these trends remain consistent with limited labeled data and with noisy shape regions.
Finally, we applied NGSP to a set of `in the wild' CAD shapes and validated that it produced better semantic decompositions than alternative approaches with a perceptual study.

When presented with an unstructured shape that lacks a region decomposition, NGSP must rely on other methods to produce suitable regions.
Many methods that decompose shapes represented as raw sensor input (e.g. point clouds) into primitive parts do so at too coarse a granularity for fine-grained segmentation~\cite{tulsiani2017learning,AdaptiveHierarchicalCuboidAbstraction,deng2020cvxnet,NeuralStarDomain,NeuralParts}.
However, the input region requirements for NGSP may actually be weaker than what most of these approaches aim to produce:
as shown in Sections~\ref{sec:res_corrupt} and \ref{sec:acd_exp}, NGSP offers advantages even when the input regions poorly approximate the target part instances. 
Developing unsupervised methods for producing such `instance over-segmentations' is a good direction for future work.

Looking forward, we believe that NGSP's framing of 3D shape semantic segmentation as approximate inference in a probabilistic model suggests a vision for how this task could be scaled beyond carefully-curated research datasets to `in-the-wild' scenarios.
In the future, we plan to design likelihood terms that cannot be easily accommodated by end-to-end approaches; these could include hard-to-differentiate terms that consider functional part relationships such as adjacency, symmetry, or physical support (e.g. the chair base should physically support the chair seat).
These terms could potentially be provided by a person via explicit rules, either in advance or with a human-in-the-loop system.
Paradigms that allow integration of such symbolic rules with data-driven models could be a key step towards producing high-quality semantic segmentations in few-shot or zero-shot scenarios.

\section*{Acknowledgments}
We would like to thank the participants in our user study for their contribution to our research.
We would also like to thank the anonymous reviewers for their helpful suggestions. 
Renderings of part cuboids and point clouds were produced using the Blender Cycles renderer. This work was funded in parts by NSF award \#1941808 and a Brown University Presidential Fellowship. Daniel Ritchie is an advisor to Geopipe and owns equity in the company. Geopipe is a start-up that is developing 3D technology to build immersive virtual copies of the real world with applications in various fields, including games and architecture

{\small
\bibliographystyle{ieee_fullname}
\bibliography{main}
}

\appendix

\section{Data Details}

\subsection{Data Preprocessing}
\label{sec:filter}
Ground-truth shape regions for instances in our datasets are created from PartNet \cite{PartNet}.
Specifically, we create a shape region for each connected mesh-component of a given shape in PartNet.
This guarantees the region decomposition is at minimum an instance segmentation, as each leaf part instance must correspond with at least 1 connected mesh-component, but it many cases it creates an instance over-segmentation, as a single leaf part instance will be represented by more than 1 mesh-component. 
To determine the semantic label of a region, we find the terminal label in the shape grammar that is a parent of the label assigned to the shape region by PartNet. 
Note that there will only ever be one such parent (because of the unique path assumption from Section 3).
There are some shape instances where there are no parent labels that meet this criteria, in which case we do not include the shape in our datasets. 
Details of the shape grammars we use can be found in Section \ref{sec:add_qual}.
Additionally we filter out any shape instances with very small part regions that would be hard to reason over with point clouds of size 10000 (the number of points used by all point cloud consuming networks we consider). 
If any shape region has an area that makes up less than 0.1\% of the total shape area, the entire shape instance is ignored.

\subsection{Creating Dataset Splits}

We used shapes from the chair, lamp, table, vase, knife, and storage furniture categories of PartNet. These were all the categories that had at least 250 valid shape instances (as defined by criteria in Section \ref{sec:filter}) and had shape grammar definitions at a fine granularity. For each of these datasets, we created train/val/test splits so that no set had more than 400 shapes or less than 50 shapes. Additionally, when making the splits we employed a greedy strategy where we tried to keep each terminal label of the grammar evenly represented (both across and within sets), whenever possible. Exact numbers of shapes in each split, for each category, can be found in Table \ref{tab:data_splits}. For all experiments, validation and test sets remain constant. In the experiment where the number of training set examples is varied, each training set at a smaller size is a proper subset of the corresponding larger training set.

\begin{table}[]
    \small
    \centering    
    \begin{tabular}{@{}ccccccc@{}}
        \toprule
        \textbf{Set} & \textbf{Chair} & \textbf{Lamp} & \textbf{Table} & \textbf{Storage} & \textbf{Vase}  &\textbf{Knife}  \\
        \midrule
        Train      & 400 & 400 & 400 & 400 & 400 & 239 \\
        Validation & 400 & 157 & 400 & 55 & 54 & 50 \\
        Test       & 400 & 157 & 400 & 55 & 54 & 50 \\
        \bottomrule
    \end{tabular}
    \caption{The number of shapes in the train, validation, and test set splits for each category.}
    \label{tab:data_splits}
\end{table}

\subsection{Region Corruption}

In Section 4.5 of the main paper, we describe an experiment where we analyze the robustness of NGSP to corruptions of the input regions. 
We experiment with two levels of corruptions, the 2X paradigm, where each region is split into 2 regions, and the 4X paradigm, where each region is split into 4 regions.
The corruptions for the 2X paradigm are produced in the following manner. 
For each region, a random mesh face, $f_0$, from that region is sampled. 
Then, treating each face as the average of its vertices, the furthest mesh face from $f_0$, $f_1$, is found. 
We then find the furthest mesh face from $f_1$ within the region, $f_2$. For all other faces within the region, we record their distance from both $f_1$ and $f_2$, and assign each face to the cluster it is closest to. 
Regions containing a single mesh face are not further split.
To generate the 4X splits, the 2X split procedure is applied twice in succession. 

\section{Training Details}

\subsection{Training Hyperparameters}

The geometry, layout, and guide networks are variants of PointNet++'s written in PyTorch \cite{qi2017pointnet++, pytorchpointnet++, paszke2017automatic}. 
All PointNet++'s use 3 set abstraction layers with 1024, 256, 64 grouping points, 0.1, 0.2, 0.4 radius size and 64, 128, 256 feature size respectively, with the global pooling step done at a feature size of 1024. ReLU activations are used throughout.
All multilayer perceptron (MLP) heads for per-shape or per-point classification use 2 hidden layers with dimensions of 256 and 64 with leaky ReLU activations (slope 0.02). 
We train with batch size of 16, without any batch normalization, but with dropout in the MLPs of 0.4 within the guide network and 0.2 for the geometry and layout networks.
For the guide network, batch members are randomly sampled from the data.
For the geometry and layout network, each batch contains one positive example, and the rest of the batch contains negative examples mined with respect to the positive one. The binary classification loss is computed with a mean operation independently for the positive and negative examples, and then summed together. 
We convert mesh data into 100000 points sampled uniformly from the surface of each shape, in order to be able to reason about regions of small, fine-grained parts.  
The guide network uses point clouds with 10000 points while the geometry and layout networks uses 4096 points.
If there are less than the expected number of points corresponding to the union of regions needed while querying the geometry or layout networks, we repeat points in order to facilitate batch training and evaluation.
All networks receive data augmentation in the form of random non-uniform scales and point-wise perturbations. 

The region network uses a graph neural network architecture that operates over a graph of shape regions;
we describe the graph generation process in Section \ref{sec:graph_creation}, that constructs the graph topology and initializes the per-node and per-edge features. 
The region network performs 4 cycles of gated graph convolution \cite{dwivedi2020benchmarkgnns, bresson2017residual}, with residual node connections, a batch size of 16, and no batch normalization,
After the graph convolutions, a global readout max-pooling operation is applied across all node features within each graph, so that there is a single feature representation for each graph.
For the label region network, a linear layer takes this feature representation and predicts a probability distribution over terminals in the grammar. 
For the area region network, a linear layer takes this feature representation and predicts a single scalar value. 

\subsection{Semantic Label Negative Sampling Strategies}
\label{sec:strat}
As mentioned in the main paper, positive examples for the geometry and layout networks are sourced directly from the ground-truth label assignments to regions of the input shapes. 
Negative examples are sourced by finding label assignments different from the canonical version, that would change the assigned regions, or child labels of the assigned regions, for the label of interest.
However, negative examples that are too similar to their corresponding positive examples are not used. 
To check this, we measure the percentage of region area that is unchanged between the positive and negative example, if this percentage is above .95, then we ignore that negative example. We employ a set of negative sampling strategies to find suitable `incorrect' label assignments, detailed as follows:

\paragraph{Label Assignment Perturbations}

One way to source negative examples is to find `incorrect' label assignments that are deviations from the original label assignment to the entire shape. We do this by creating 9999 perturbed label assignments for each shape, where 100 have 1 label change, 200 and 2 label changes, 300 have 3 label changes, 400 have 4 label changes, 500 have 5 label changes, 500 have 10\% label changes, 1000 have 20\% label changes, 1500 have 30\% label changes, 2000 have 40\% label changes, 2500 have 50\% label changes and 999 have all labels changed. 
When sampling label changes, we make it more likely that labels will be switched to other labels in the grammar they are closer to in the hierarchy of labels implied by the grammar.

\paragraph{Adding Regions}

In this method to source negative examples for a given shape and label, we first identify all regions that are assigned to this label, and all regions that are not assigned to this label, under the canonical labeling. Then some regions (randomly sampled) that are not assigned to this label are switched to be assigned to the label of interest.

\paragraph{Removing Regions}

In this method to source negative examples for a given shape and label, we first identify all regions that are assigned to this label under the canonical labeling. Then some regions (randomly sampled) that are assigned to this label are removed and are no longer assigned to the label of interest.

\paragraph{Using Regions from different Parts}

In this method to source negative examples for a given shape and label, we first identify all regions that are not assigned to this label under the canonical labeling. Then a subset of these regions (randomly sampled) are assigned to the label of interest (with all other regions in the shape unassigned to this label).

\paragraph{Using Regions from a different Shape}

In this method to source negative examples for a given shape and label, we first find all other shapes in the dataset where the label of interest was not seen, and sample one such shape. Then a subset of regions (randomly sampled) from this sampled shape is assigned to the label as a negative example.

\paragraph{Changing the Child Label of Regions}

This method to source negative examples for a given shape and label is only used by the layout networks. No regions of the input shape are changed, instead for each region there is a 50\% chance to change the assigned child label of the region to a different random value (consistent with the available options defined by the grammar). 

For all other negative sampling strategies, when negative examples are generated for the layout network and a new region is added, a random child label is assigned in the same fashion.

\subsection{Differentiating Geometry and Layout Networks}

The differences between the geometry and layout networks come from the input features, the labels in the shape grammar they cover, and the types of negative examples they learn from.
Both types of networks principally operate over points that come from regions assigned to the label of interest. 
The layout network additionally receives the assigned child label of each point with respect to the corresponding label as a onehot vector concatenated to the XYZ position of each point.
As the terminal labels of the shape grammar have no children labels, they are not assigned any layout networks.
As the root label of the shape grammar always encompasses all shape regions, there is no geometry network assigned to it.
The types of negative examples seen by each network are sampled at different rates.
For the geometry network, the sampling probabilities for each negative example are (corresponding to the strategies listed in \ref{sec:strat}): 50\% label assignment perturbations, 15\% adding regions, 15\% removing regions, 15\% using regions from different parts, 5\% using regions from a different shape and 0\% changing the child label of regions. 
For the layout network, the sampling probabilities for each negative example are (corresponding to the strategies listed in \ref{sec:strat}): 50\% label assignment perturbations, 7.5\% adding regions, 7.5\% removing regions, 7.5\% using regions from different parts, 2.5\% using regions from a different shape and 25\% changing the child label of regions. 
\subsection{Region Graph Creation}
\label{sec:graph_creation}

The region network takes as input a collection of shape regions represented as a graph.
The nodes of this graph correspond to shape regions, and the edges are created such that the graph is fully connected.
To populate the initial node and edge features we use point cloud auto-encoders.
The point cloud auto-encoders are trained on a collection of chair shapes we gather from PartNet; we make use of their shape region decompositions but do not use their label information. 
For the node feature, we train a PointNet++ to consume a 1024 dimensional point cloud sampled from one shape region, project it into a 64 dimensional latent space, and then decode the 64 dimensional vector with a 3-layer MLP into a 1024 x 3 vector; encouraging the input and output point clouds to match with a Chamfer distance loss.
For the edge feature, we train a point cloud auto-encoder to consume two 1024 dimensional point clouds sampled from two shape regions. The point clouds are distinguished from one another with a one hot encoding appended to each XYZ position. These point clouds are concatenated together to form a 2048 x 5 input vector. 
A PointNet++ module consumes this input and projects it into a 64 dimensional latent space.
This 64 dimensional vector is then run through a 3-layer MLP to form a 2048 x 3 vector. 
We interpret the first 1024 of these points to correspond to the first region and the last 1024 of these points to correspond to the second region, and encourage each decoded point cloud to match its target with a Chamfer distance loss.
For both paradigms, all shape regions are centered and scaled to lie within the unit sphere, and training is done with the Adam optimizer, a learning rate of 0.0001, and a batch size of 32. 
Both the region point cloud auto-encoder and the paired-region point cloud auto-encoder are pretrained and frozen; they are used across categories and labeled data training set sizes. 
Finally, when creating region graphs for the region network, per-node features are created by running the region through the region point cloud auto-encoder, and per-edge features are created by running pairs of regions through the paired-region point cloud auto-encoder.
The position and scale of each region are also concatenated onto each per-node feature.

\section{Perceptual Study}

\begin{figure*}[t!]
    \centering
    \includegraphics[{width=\linewidth}]{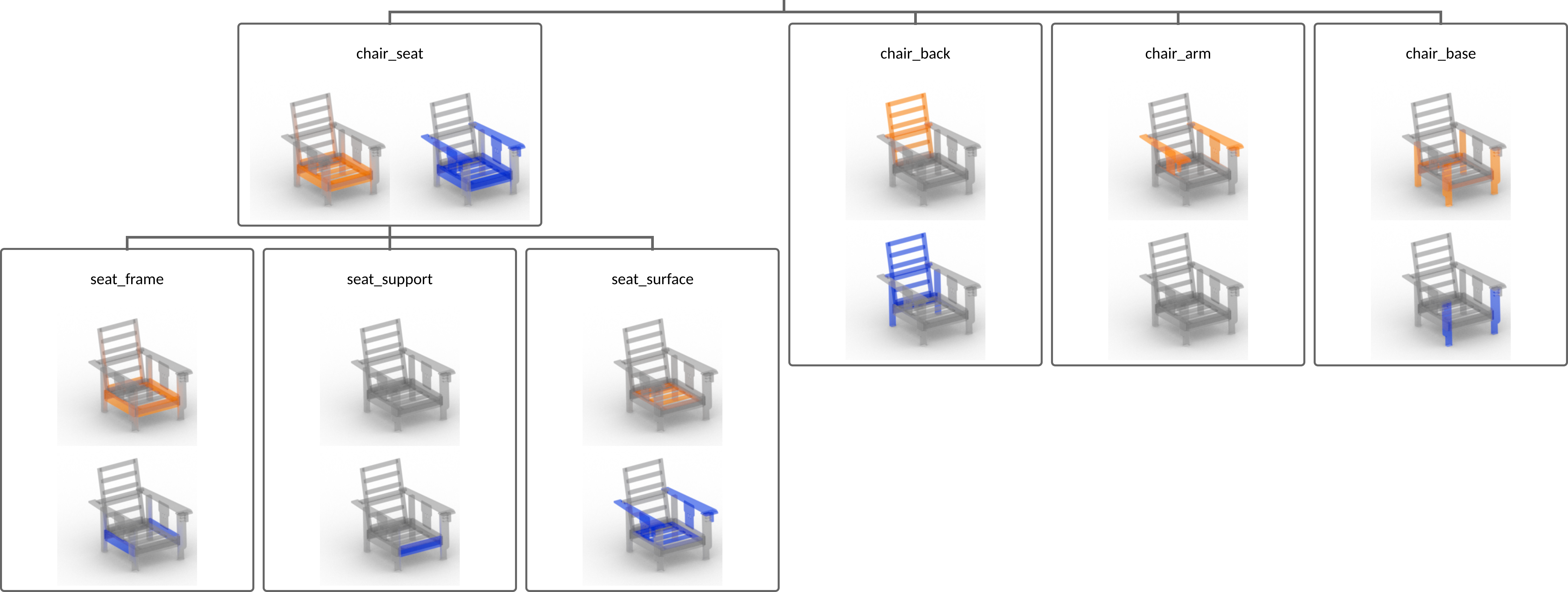}
    \caption{Example comparison from our perceptual study visualizing two label assignments to shape regions sourced from a ShapeNet mesh \cite{chang2015shapenet}. The orange labeling is from NGSP and the blue labeling is from LHSS. The supplemental includes additional examples.}
    \label{fig:qual_perceptual}
\end{figure*}

As described in Section 4.6 of the main paper, we ran a perceptual user-study to determine semantic segmentation performance for `in the wild` shape instances that lacked ground-truth label annotations.
We recruited 12 college students to participate in our study. 
Each participant was shown a sequence of 46 shape segmentation examples.
Each example compared the shape segmentations produced by two different methods (either NGSP and PartNet or NGSP and LHSS).
We visualized each labeling by expanding the label hierarchy of the grammar. 
For each label of the grammar, when the two label assignments agreed on which parts were assigned to that label, that label was colored purple.
When the two methods differed on which parts were assigned to that label, each method's parts were depicted side-by-side and given different colors (orange or blue).
Figure \ref{fig:qual_perceptual} shows an example prompt, where the orange labeling is predicted by NGSP and the blue labeling is predicted by LHSS. 
For each example, the participants were asked to pick the label assignment (orange or blue) that better matched the given shape; we reported quantitative results of this study in Table 4 of the main paper.
For this quantitative analysis we included all participants who spent between 5 minutes and 30 minutes on this task; excluding 2 outliers who took 2 minutes and 4 hours respectively to complete the survey. 

\begin{figure*}[]
      \vspace{-4.0mm}
    \centering
    \setlength{\tabcolsep}{1pt}
    \begin{tabular}{ccccc}
        Input Regions & PartNet & LHSS & NGSP & GT \\
        \includegraphics[trim={12cm 10cm 9cm 10cm},clip,{width=.2\linewidth}]{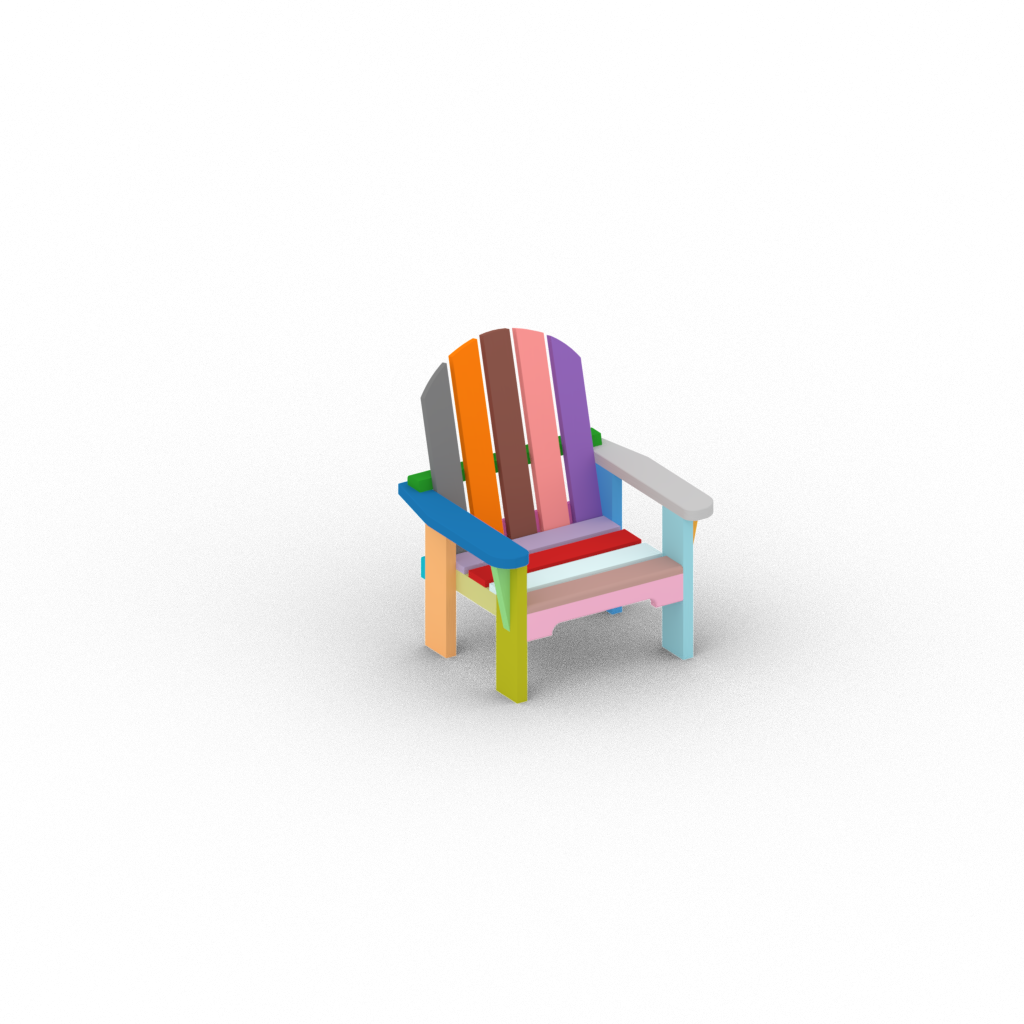} &
        \includegraphics[trim={11cm 10cm 9cm 10cm},clip,{width=.2\linewidth}]{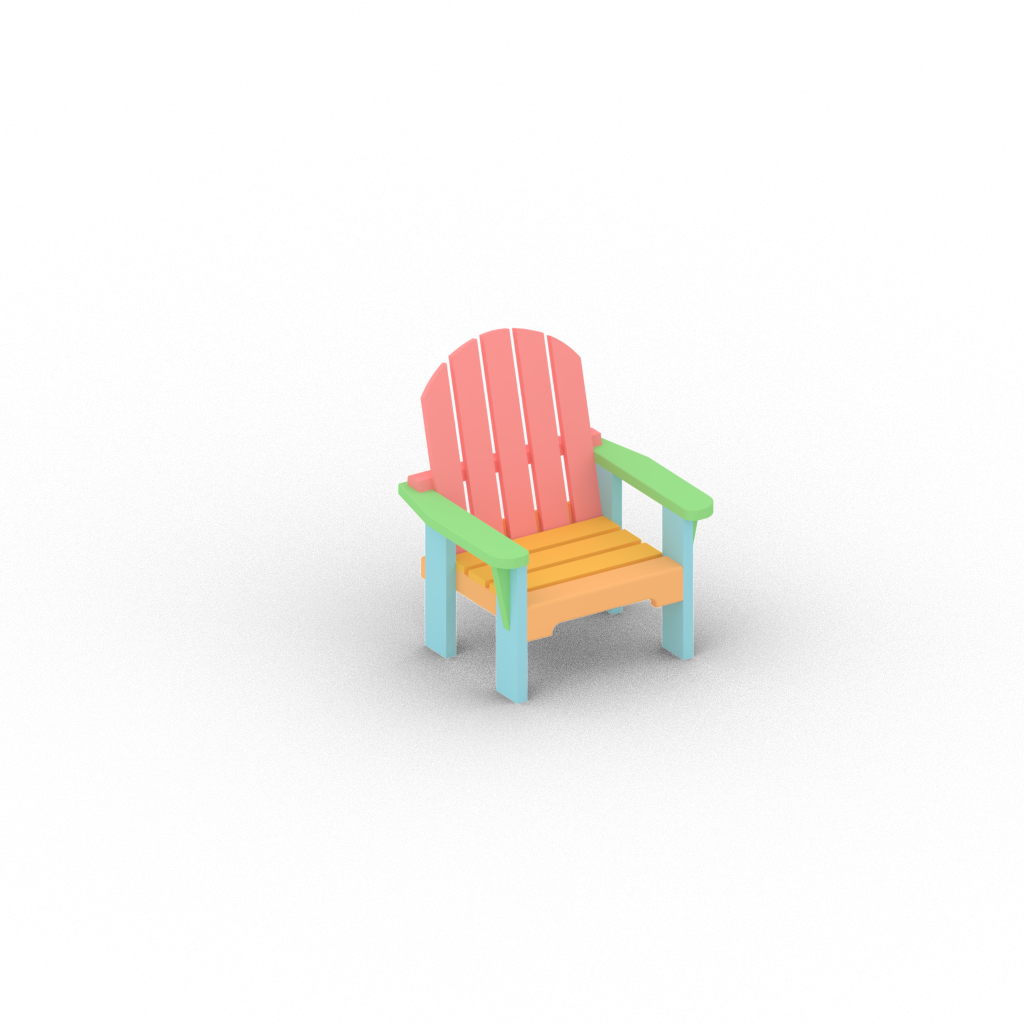} &
        \includegraphics[trim={11cm 10cm 9cm 10cm},clip,{width=.2\linewidth}]{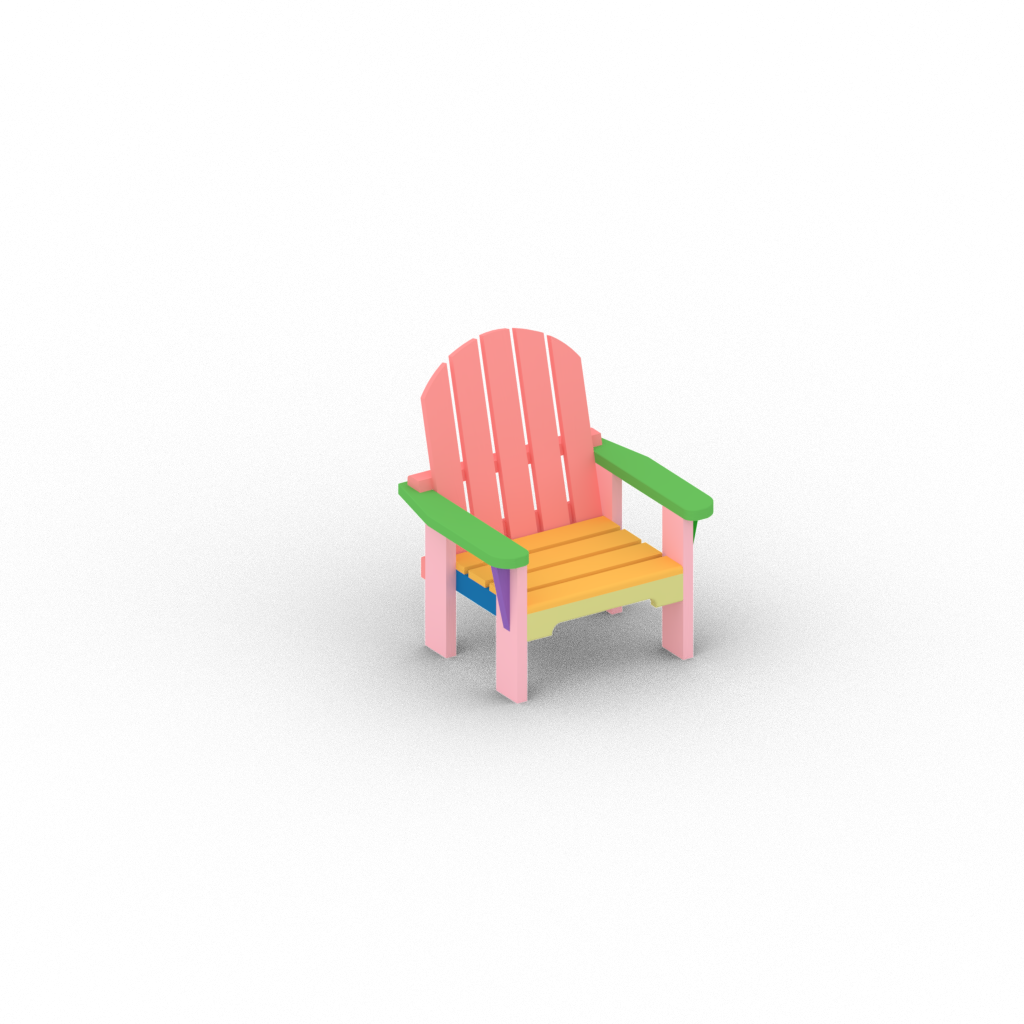} &
        \includegraphics[trim={11cm 10cm 9cm 10cm},clip,{width=.2\linewidth}]{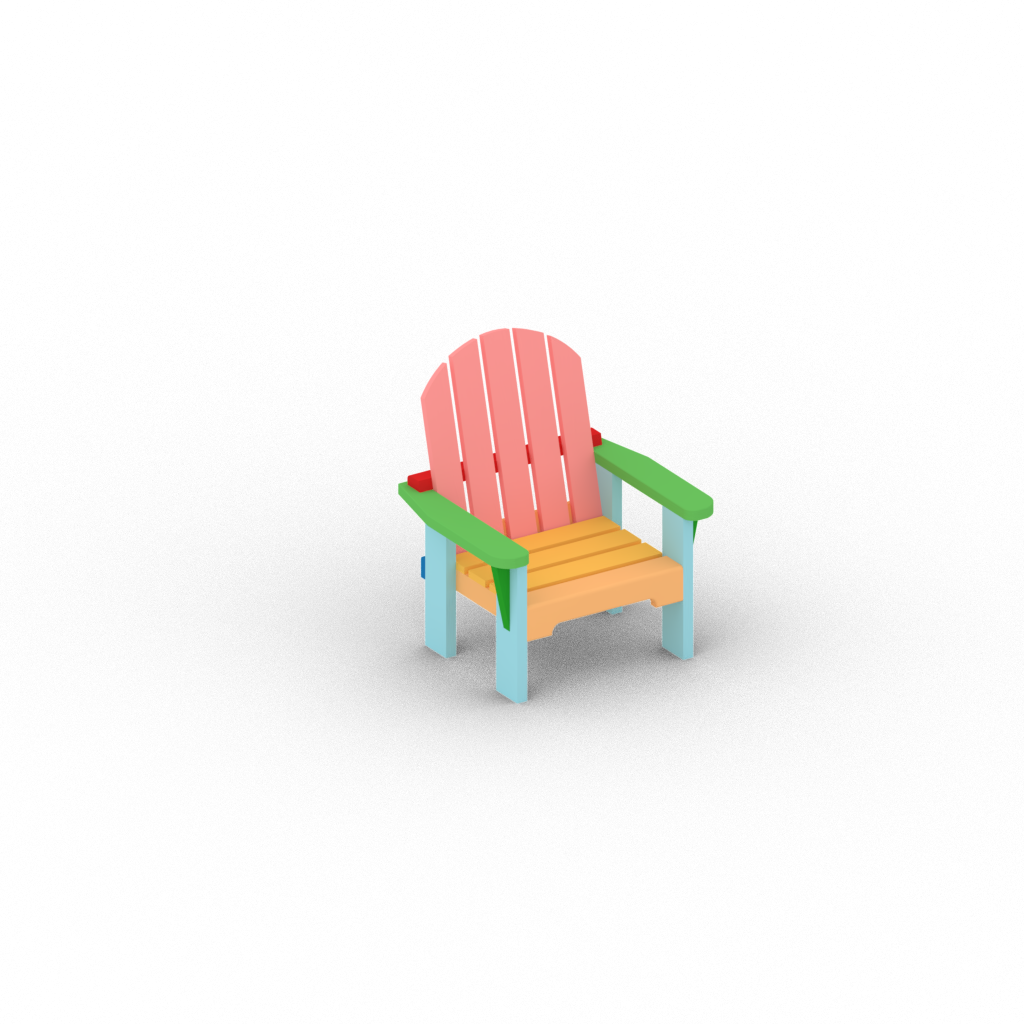} &
        \includegraphics[trim={11cm 10cm 9cm 10cm},clip,{width=.2\linewidth}]{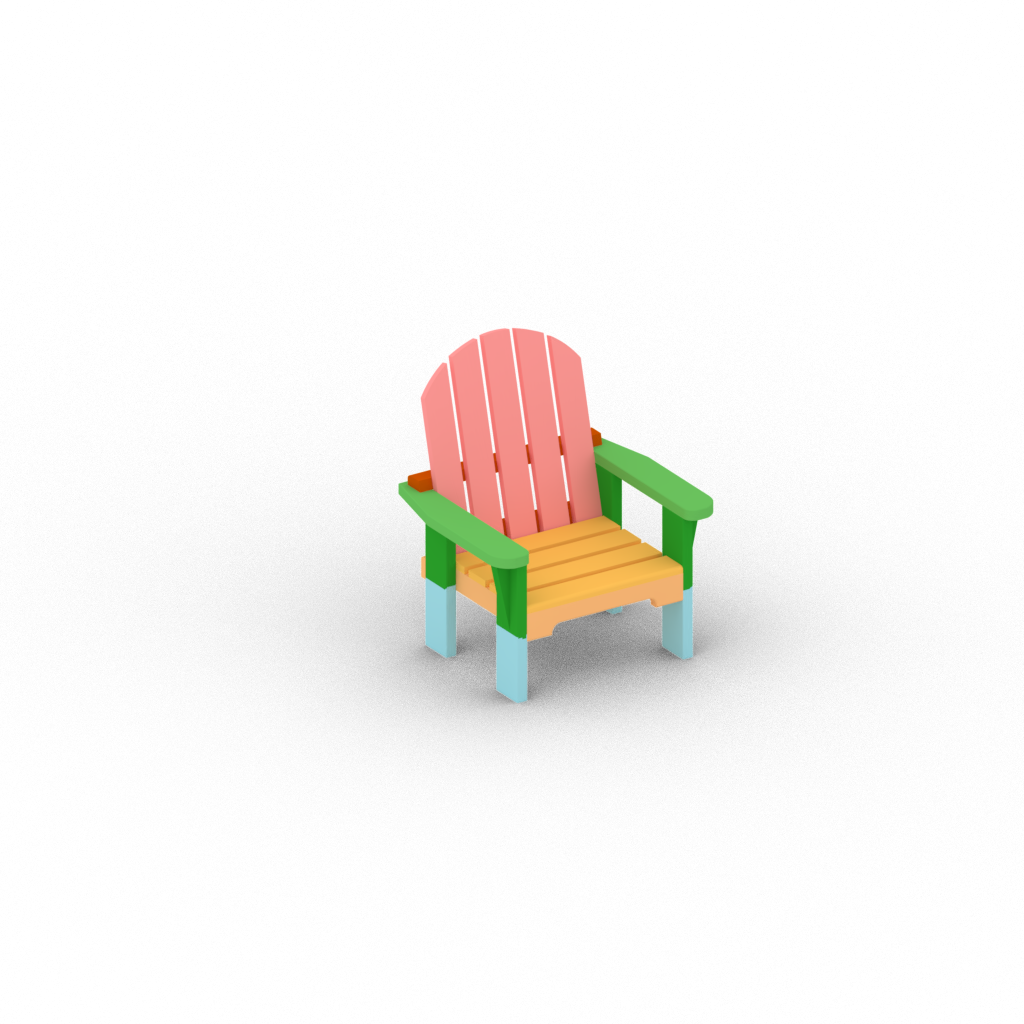} \\

        \includegraphics[trim={11cm 10cm 9cm 10cm},clip,{width=.2\linewidth}]{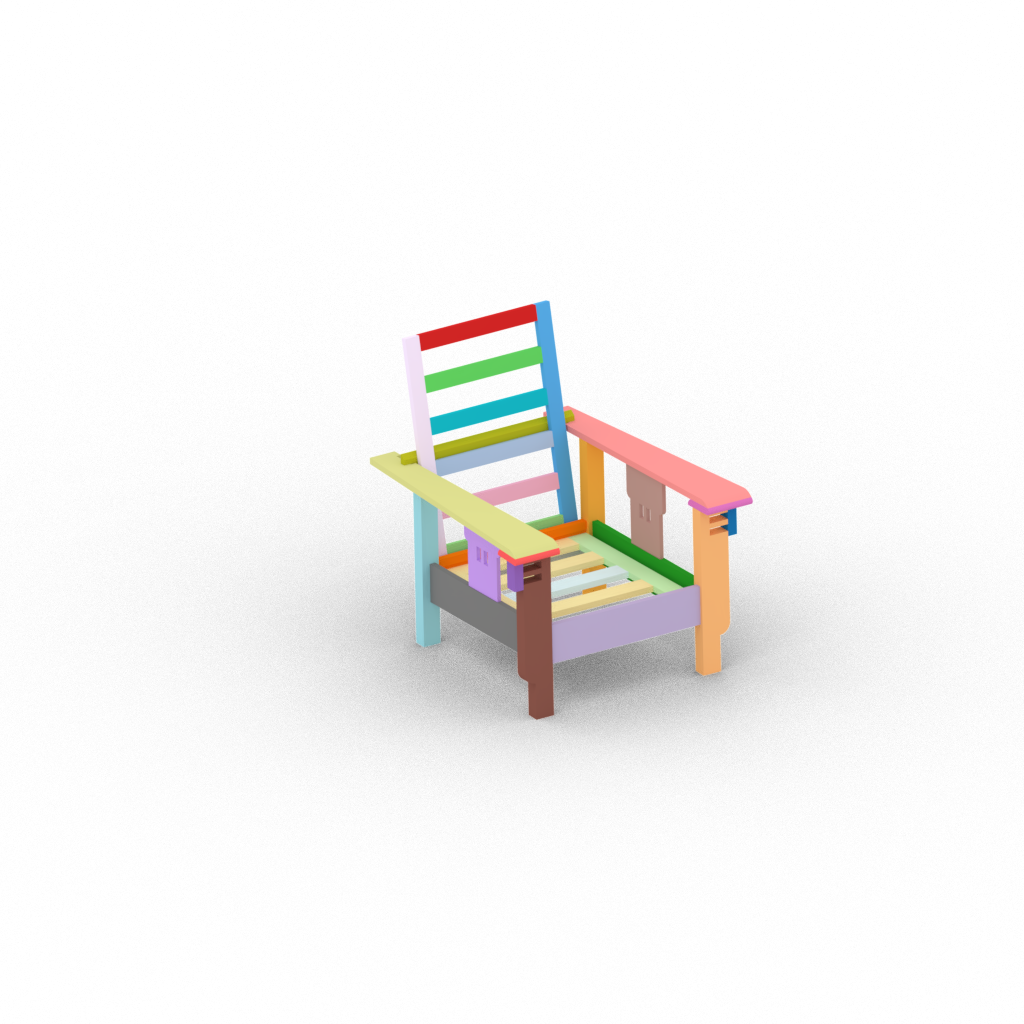} &
        \includegraphics[trim={11cm 10cm 9cm 10cm},clip,{width=.2\linewidth}]{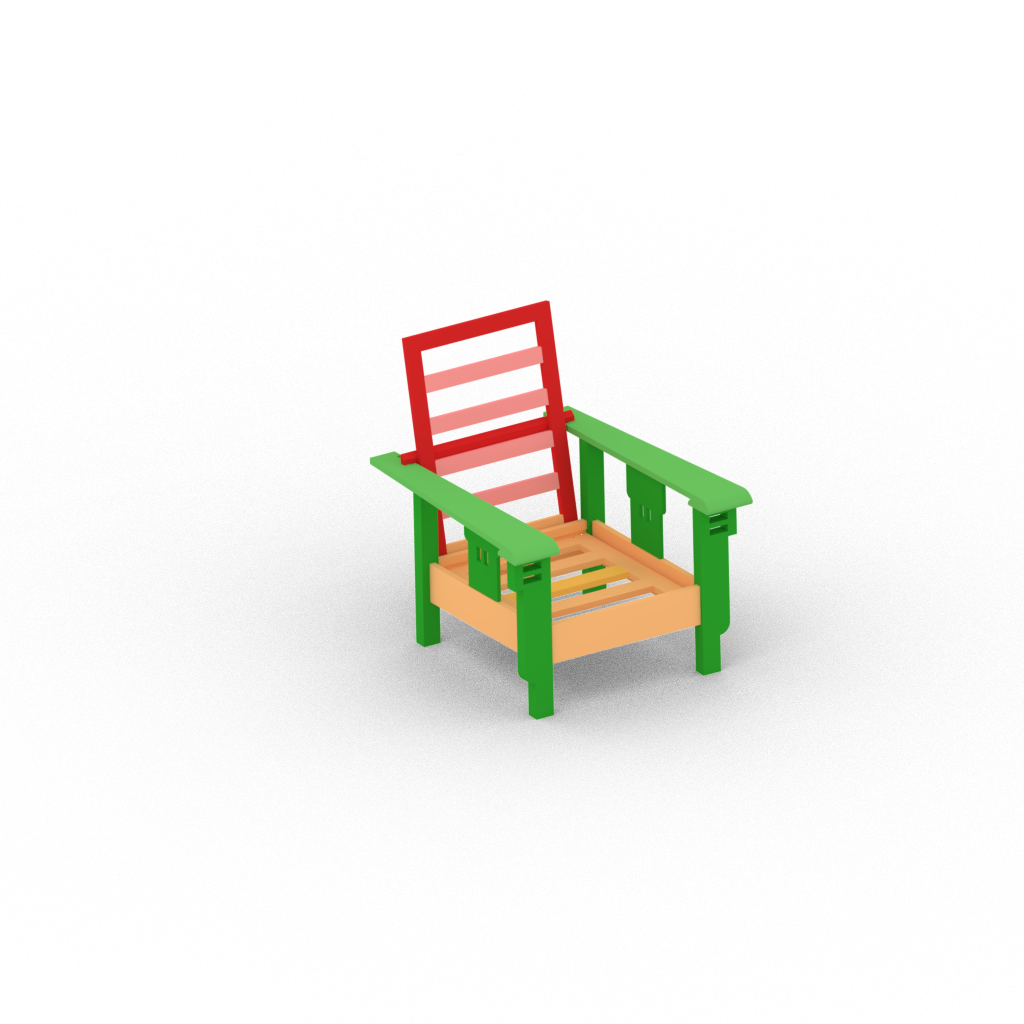} &
        \includegraphics[trim={11cm 10cm 9cm 10cm},clip,{width=.2\linewidth}]{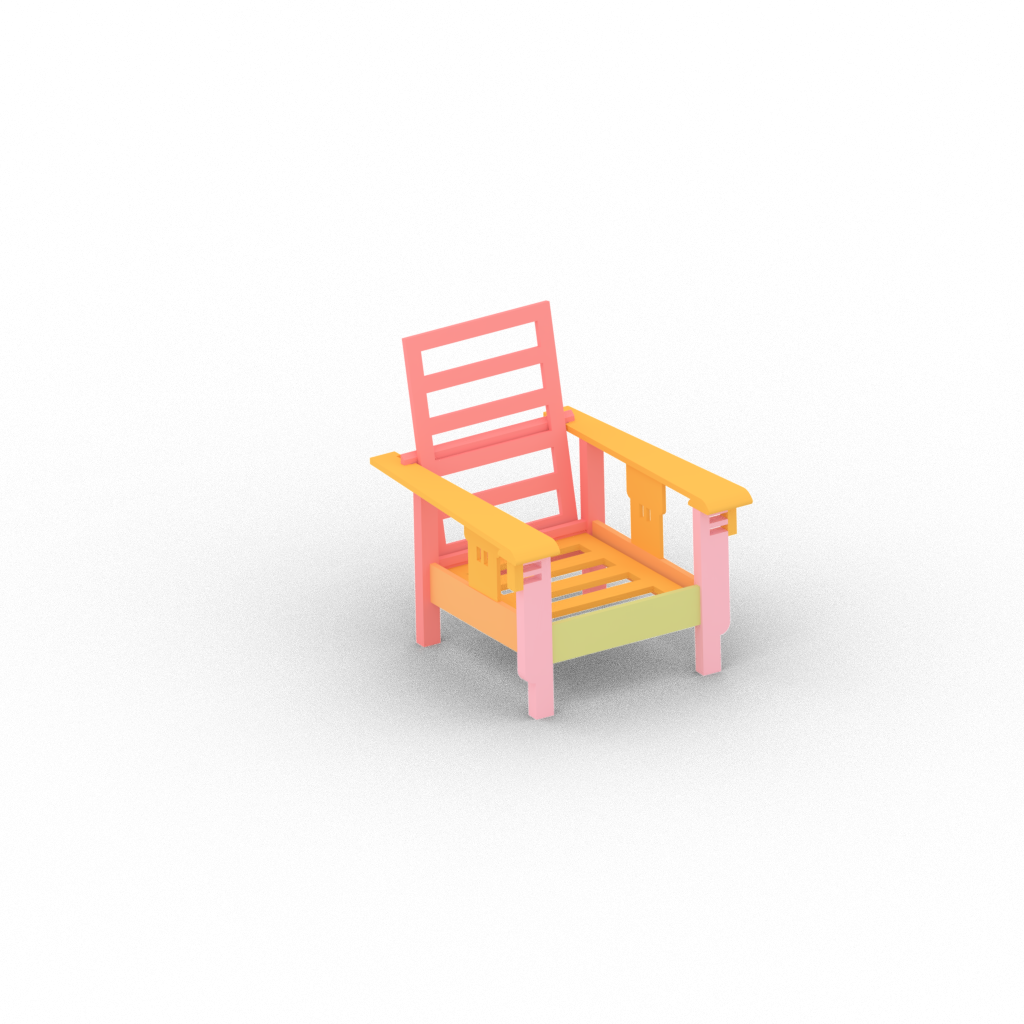} &
        \includegraphics[trim={11cm 10cm 9cm 10cm},clip,{width=.2\linewidth}]{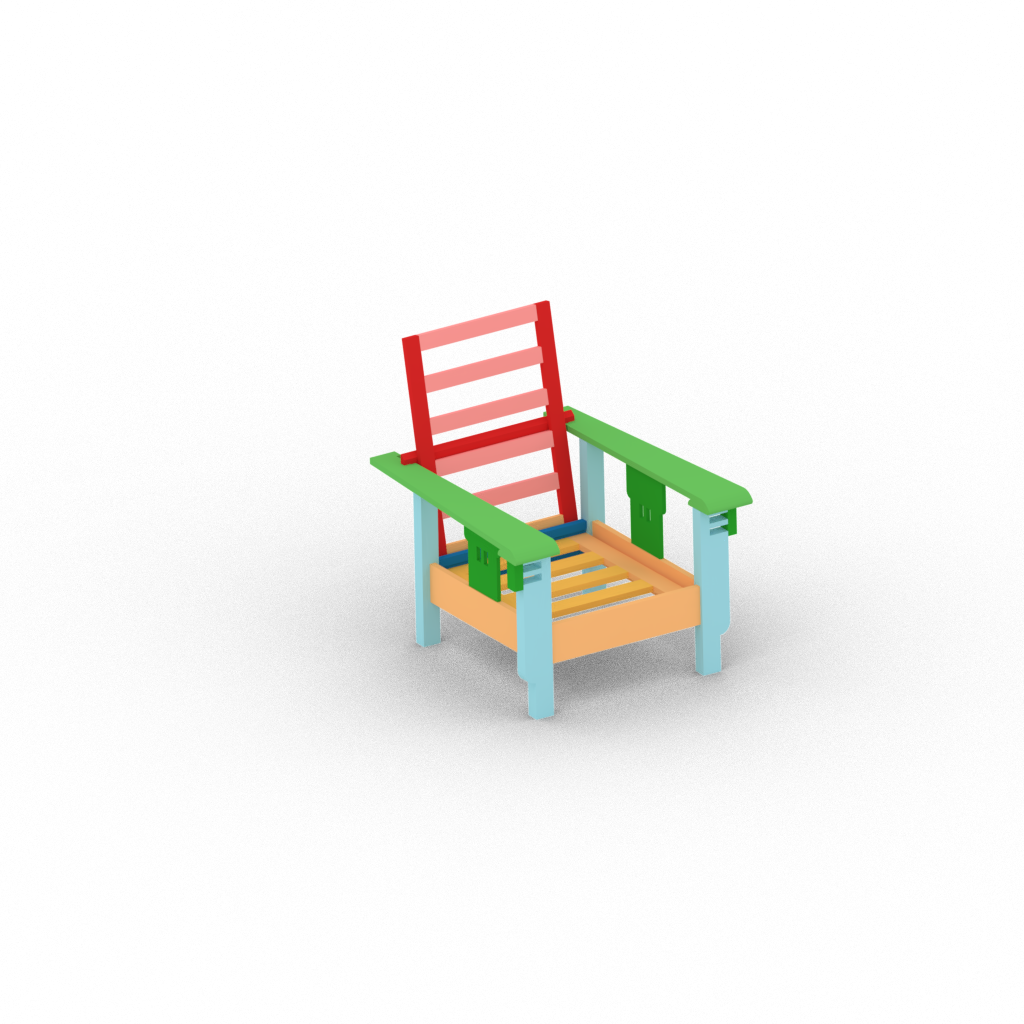} &
        \includegraphics[trim={11cm 10cm 9cm 10cm},clip,{width=.2\linewidth}]{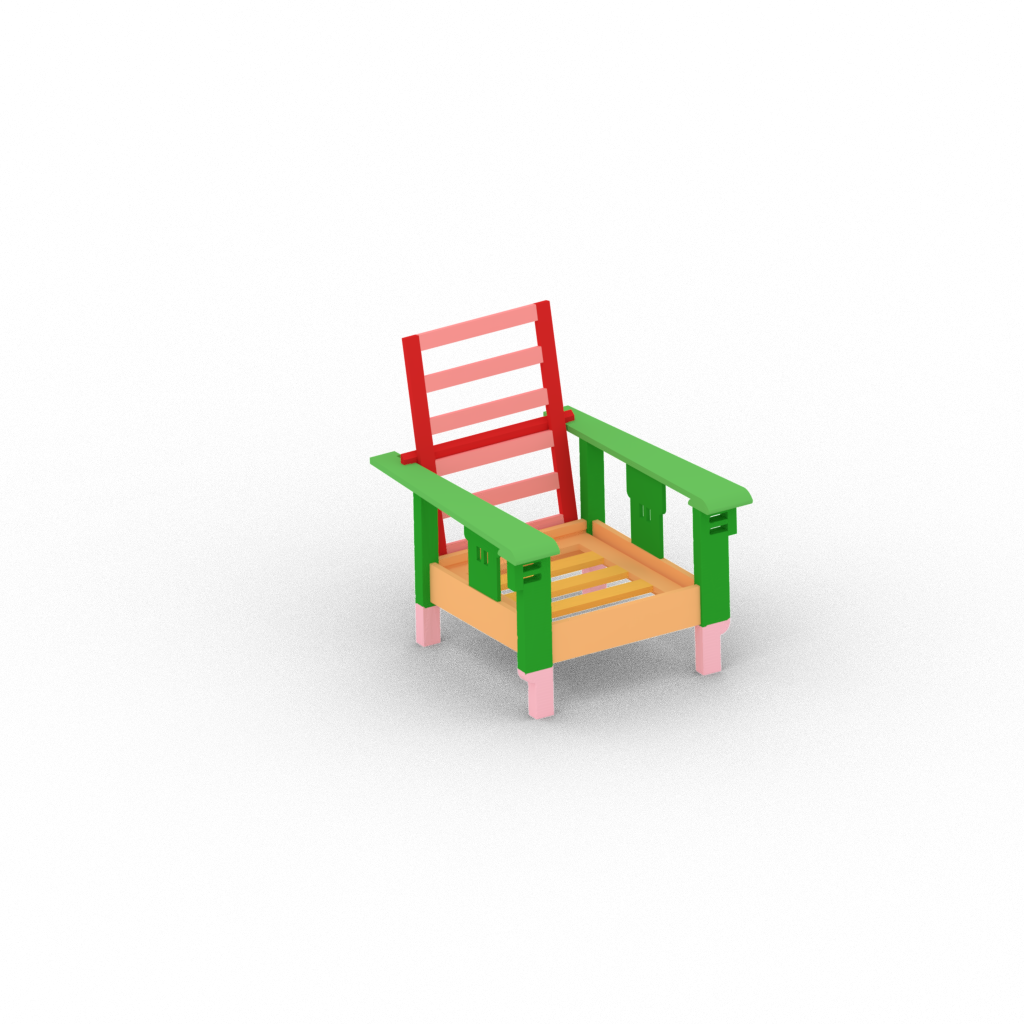} \\
    \end{tabular}
    \vspace{-3.0mm}
    \caption{Additional qualitative results for ‘in the wild’ shapes.}
    \label{fig:rebut_qual}
\end{figure*}

As described in the main paper, we sourced the `in the wild' shapes from the ShapeNet dataset \cite{chang2015shapenet}. 
We used shape instances from the chair category, and collected a set of 26 meshes whose given connected-components roughly corresponded to part-instance over-segmentations.
For each of these 26 meshes, we produced 2 potential label assignment comparisons (for comparisons against both PartNet and LHSS). 
The choice of which method should be colored blue or orange was randomized for every rendering.
To not overwhelm each participant, instead of expanding the entire chair grammar hierarchy, for each example we always show all the children of the root node (chair back, base, seat, arm, head, footrest), and we randomly choose to show the full expansion of exactly one child.
Depending on the given label assignments, some root children expansion views did not present a substantial qualitative difference between the two semantic segmentation methods (either because of very small and hard to perceive shape regions or because the chosen label assignments were the same for the expanded nodes); we manually removed such expansions from the experiment, reducing the number of examples each participant was asked to make judgement on from 52 -> 46. 
In the supplemental zip-file, we include all of the comparison renders that participants were shown in the experiment. 
In Figure \ref{fig:rebut_qual} we include qualitative comparisons of predicted labelings from different methods.

\section{Comparison Method Implementation Details}

\subsection{BAE-NET}

We follow the implementation provided by the BAE-NET author's whenever possible \cite{chen2019bae_net}.
To produce voxelizations that BAE-NET takes as input, we take the following steps. 
First we create a manifold version of the mesh \cite{huang2018robust}.
Then we compute inside-outside values for query points that lie along a grid, using the fast winding number algorithm \cite{Barill:FW:2018}.
We use logic from the BAE-NET code to turn these query point inside-outside values into a voxelization for each shape (used as input to an encoder) and paired (point, value) data used to train the implicit decoder. 

BAE-NET has two training modes: supervised and unsupervised training. 
Supervised training can only be run on shapes that come with semantic labels. 
In the original BAE-NET implementation, 3000 warm-up epochs are run over the full supervised learning set (all shapes that contain labels), then unsupervised shapes are integrated. 
After the warm-up phase, after every 4 unsupervised training updates, BAE-NET makes a supervised learning update for every shape instance in the supervised set. 
We employ this paradigm in the low-data regimes (10/40 labeled data instances).
In plentiful labeled data regimes, this strategy is prohibitively slow, so we instead make one full pass through both the unsupervised and supervised examples for each epoch after the warm-up period.

\subsection{LEL}

Our label-efficient learning variant utilizes the shape region based self-supervised training scheme proposed in \cite{selfsupacd}.
In their experiments, they source shape region decompositions from an ACD method.
In our experiments, we use shape region decompositions provided by PartNet. 
Following their method, we modify the PointNet++ used in the PartNet variant to include an additional linear layer, that consumes the last per-point feature and outputs a 128 dimensional per-point embedding.
The self-supervised loss encouraging per-point embeddings to cluster to similar parts of space is then implemented with code from the label-efficient learning paper.

\subsection {LHSS}

We follow the implementation provided by the method's authors whenever possible \cite{HierarchicalSegLabelOnline}.
We directly use their Julia code that consumes input meshes in order to generate the per-point features.
We re-implement their neural network in PyTorch, following all hyper-parameters as described in their released torch code. 
To solve the MRF formed by per-shape-region unary terms and paired terms that correlate to label distances within the grammar hierarchy, we use the alpha-expansion algorithm from the publically available GCO package \cite{10.1109/34.969114}.

\subsection{Converting Per-Atom Predictions to Per-Region Predictions}

As mentioned in the main paper, some methods make per-point predictions, e.g. they predict a semantic label for each point in the input point cloud (PartNet, BAE-NET, LEL).
For a fair comparison against NGSP and LHSS, which assign labels to shape region, for methods appended by (R) we group per-atom predictions into per-region predictions.
To do this, we first compute the probability distribution over labels for each point (e.g. send the logits through a softmax). Then we group points based on the shape regions, and for each shape region we find the region label probability distribution by average all of the per-point label probability distributions.
We then take the arg max of the region label probability distribution as the chosen label assignment for that shape region.

\subsection{Unlabeled Additional Shape Instances }

Some comparison methods (BAE-NET and LEL) are able to use shape instances that lack label annotations, but contain shape region decompositions. 
For fair comparison, we allow these methods to train on up to 1000 additional shapes where the shape region decomposition is provided, but the semantic label annotations are withheld. 
We source these unlabeled shapes from PartNet instances that do not show up in the labeled training, validation or test sets. 
For some number of labeled training data + category combinations, there are not enough shape instances in PartNet to reach 1000, in which case we use as many shapes as there are available.
Table \ref{tab:unlabel_num} contains how many unlabeled shape instances are used by BAE-NET and LEL for each number of labeled training data + category combination.

\begin{table}[]
    \footnotesize
    \centering    
    \begin{tabular}{@{}ccccccc@{}}
        \toprule
        \textbf{\# Train} & \textbf{Chair} & \textbf{Lamp} & \textbf{Table} & \textbf{Storage} & \textbf{Vase}  &\textbf{Knife}  \\
        \midrule
        10 Labeled & 1000 & 1000 & 1000 & 1000 & 1000 & 404 \\
        40 Labeled & 1000 & 1000 & 1000 & 1000 & 1000 & 374 \\
        400 Labeled & 1000 & 1000 & 1000 & 1000 & 652 & 175 \\
        \bottomrule
    \end{tabular}
    \caption{Number of additional unlabeled shape instances used by BAE-Net and LEL comparison methods, for each category, and each number of labeled training data used during training. }
    \label{tab:unlabel_num}
\end{table}

\section{Inference Run-Time}

As demonstrated, NGSP outperforms comparison methods on the task of semantic segmentation.
This performance improvement comes at the cost of an increase in the time it takes to produce semantic segmentations; NGSP does not operate in an end-to-end fashion, but rather performs approximate MAP inference.
This search is directed by the neural guide network, which proposes a constrained set of label assignments that are then considered under the full likelihood model; evaluating more proposals will result in a better MAP estimate, but incurs more computation time.
This trade-off is controlled with the hyper-parameter $k$, the number of proposals generated from the guide network. 
When $k$ is set to 1, the performance is equivalent to just using the guide network (from the ablation table main paper).
The time it takes NGSP to generate a semantic segmentation for an average chair is 0.2 seconds when $k=10$, 0.3 seconds when $k=100$, 0.8 seconds when $k=1000$, and 4 seconds when $k=10000$.
For all results, we set $k$ to 10000, allowing us to well-approximate the MAP, while keeping computational time manageable.

%\definecolor{ppt_purple}{rgb}{0.57,0.39,0.74}

\definecolor{red_l}{RGB}{255, 152, 150}
\definecolor{red_d}{RGB}{216, 37, 38}
\definecolor{red_o}{RGB}{200, 82, 0}
\definecolor{cardinal}{RGB}{177, 3, 24}
\definecolor{green_sl}{RGB}{152, 223, 138}
\definecolor{green_l}{RGB}{103, 191, 92}
\definecolor{green_d}{RGB}{44, 160, 44}
\definecolor{yellow_d}{RGB}{255, 193, 86}
\definecolor{yellow_l}{RGB}{219, 219, 141}
\definecolor{blue_l}{RGB}{158, 218, 229}
\definecolor{blue_n}{RGB}{31, 119, 180}
\definecolor{blue_a}{RGB}{39, 190, 207}
\definecolor{purple_d}{RGB}{123, 102, 210}
\definecolor{purple_p}{RGB}{148, 103, 189}
\definecolor{fush}{RGB}{220, 95, 189}
\definecolor{orange_l}{RGB}{255, 187, 120}
\definecolor{blue_la}{RGB}{158, 218, 229}

\definecolor{dark_brown}{RGB}{101, 67, 33}
\definecolor{pink}{RGB}{255, 192, 203}
\definecolor{cafe}{RGB}{161,130, 98}
\definecolor{mocha}{RGB}{190, 164, 147}
\definecolor{teal}{RGB}{0, 128, 128}
\definecolor{baby_blue}{RGB}{137, 207, 240}
\definecolor{iceberg}{RGB}{113, 166, 210}
\definecolor{turqoise}{RGB}{64, 224, 208}
\definecolor{gold}{RGB}{204, 164, 61}
\definecolor{lime}{RGB}{206, 250, 5}
\definecolor{mint}{RGB}{255, 255, 214}

\section{Additional Qualitative Results}
\label{sec:add_qual}
\begin{figure*}[]
    \centering
    \setlength{\tabcolsep}{1pt}
    \begin{tabular}{ccccccc}
        Input Regions & PartNet (R) & BAE-NET (R) & LEL (R) & LHSS & NGSP & GT
        \\
        \includegraphics[{width=.14\linewidth}]{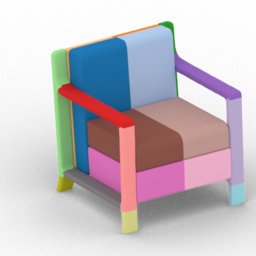} &
        \includegraphics[{width=.14\linewidth}]{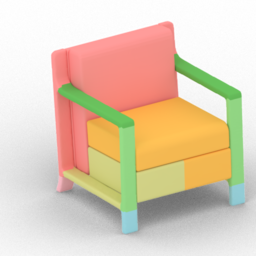} &
        \includegraphics[{width=.14\linewidth}]{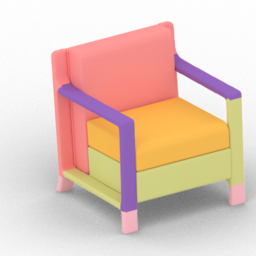} &
        \includegraphics[{width=.14\linewidth}]{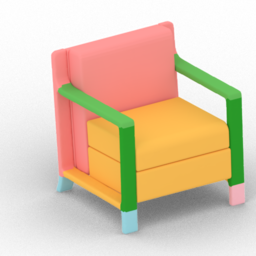} &
        \includegraphics[{width=.14\linewidth}]{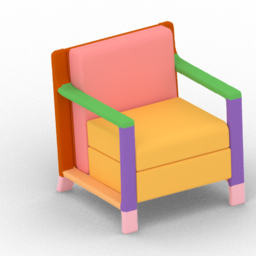} &
        \includegraphics[{width=.14\linewidth}]{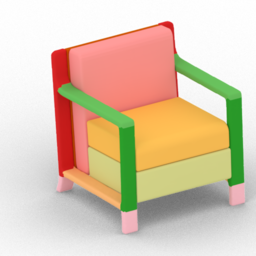} &
        \includegraphics[{width=.14\linewidth}]{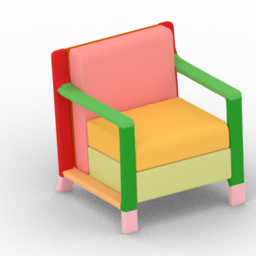} \\
        
        \includegraphics[{width=.14\linewidth}]{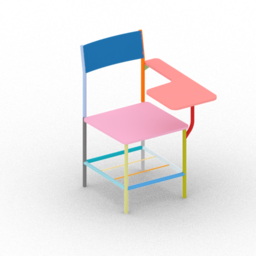} &
        \includegraphics[{width=.14\linewidth}]{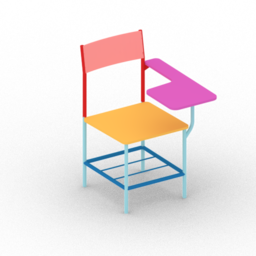} &
        \includegraphics[{width=.14\linewidth}]{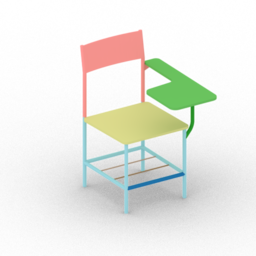} &
        \includegraphics[{width=.14\linewidth}]{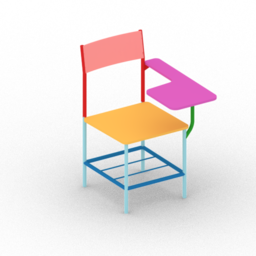} &
        \includegraphics[{width=.14\linewidth}]{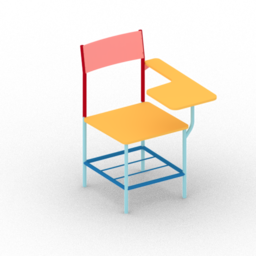} &
        \includegraphics[{width=.14\linewidth}]{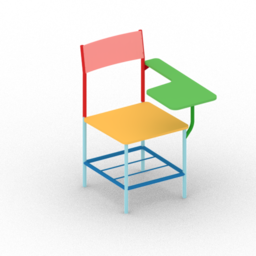} &
        \includegraphics[{width=.14\linewidth}]{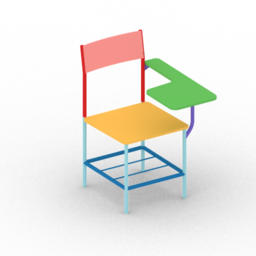} \\
        
        \includegraphics[{width=.14\linewidth}]{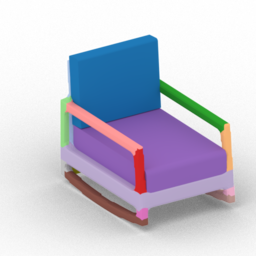} &
        \includegraphics[{width=.14\linewidth}]{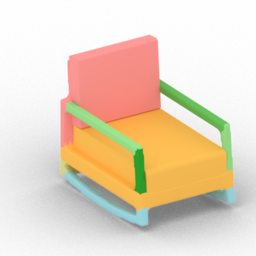} &
        \includegraphics[{width=.14\linewidth}]{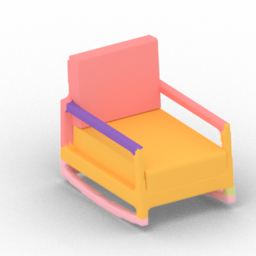} &
        \includegraphics[{width=.14\linewidth}]{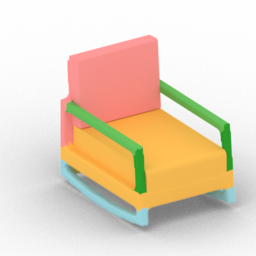} &
        \includegraphics[{width=.14\linewidth}]{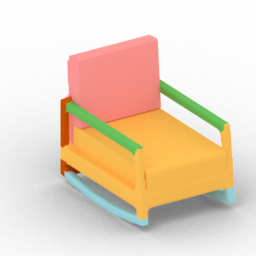} &
        \includegraphics[{width=.14\linewidth}]{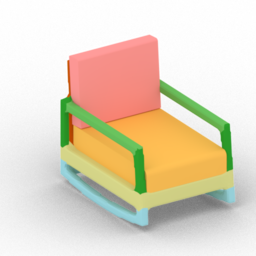} &
        \includegraphics[{width=.14\linewidth}]{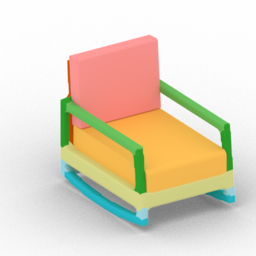} \\
    \end{tabular}
    \begin{tabular}{|l|}

\hline
\\
Chair $\xrightarrow{}$ arm; back; base; head; seat; footrest \\
Arm $\xrightarrow{}$ \colorbox{purple_d}{connector}; \colorbox{green_sl}{holistic frame}; \colorbox{green_l}{horizontal bar}; \colorbox{green_d}{near vertical bar}; sofa style; \colorbox{fush}{writing table} \\
Back $\xrightarrow{}$ \colorbox{cardinal}{connector}; \colorbox{red_d}{frame}; \colorbox{red_o}{support}; \colorbox{red_l}{surface} \\
Base $\xrightarrow{}$ foot base; pedestal base; regular leg base; star leg base \\
Head $\xrightarrow{}$ connector; headrest \\
Seat $\xrightarrow{}$ \colorbox{orange_l}{frame}; \colorbox{yellow_l}{support}; \colorbox{yellow_d}{surface} \\
Footrest $\xrightarrow{}$ base; footrest seat \\
Foot Base $\xrightarrow{}$ \colorbox{pink}{foot} \\
Pedestal Base $\xrightarrow{}$ central support; pedestal \\
Regular Leg Base $\xrightarrow{}$ \colorbox{blue_n}{bar stretcher}; foot; \colorbox{blue_l}{leg}; \colorbox{blue_a}{rocker}; \colorbox{blue_la}{runner} \\
Star Leg Base $\xrightarrow{}$ \colorbox{cafe}{central support}; mechanical control; \colorbox{mocha}{star leg set} \\
Footrest Seat $\xrightarrow{}$ support ; surface \\
\\
\hline
\end{tabular}
    
    \caption{ Qualitative Results and Grammar for the Chair category. }
    \label{fig:chair_qual}
\end{figure*}

\begin{figure*}[]
    \centering
    \setlength{\tabcolsep}{1pt}
    \begin{tabular}{ccccccc}
        Input Regions & PartNet (R) & BAE-NET (R) & LEL (R) & LHSS & NGSP & GT
        \\
        \includegraphics[{width=.14\linewidth}]{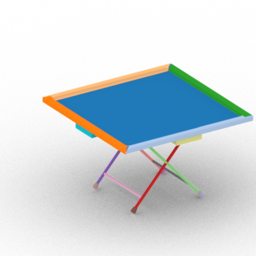} &
        \includegraphics[{width=.14\linewidth}]{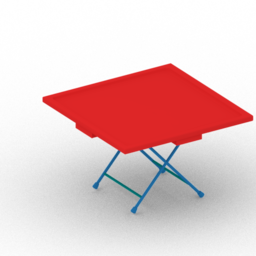} &
        \includegraphics[{width=.14\linewidth}]{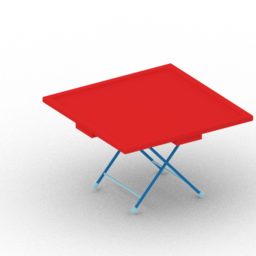} &
        \includegraphics[{width=.14\linewidth}]{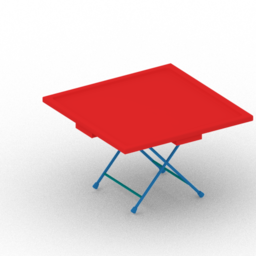} &
        \includegraphics[{width=.14\linewidth}]{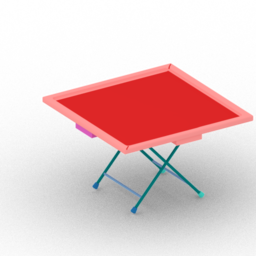} &
        \includegraphics[{width=.14\linewidth}]{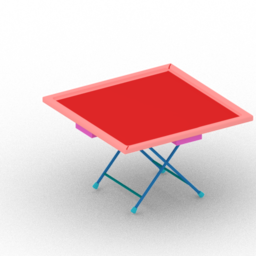} &
        \includegraphics[{width=.14\linewidth}]{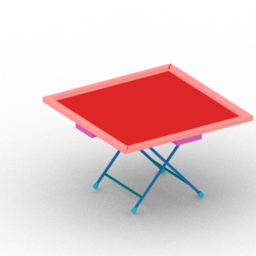} \\
        
        \includegraphics[{width=.14\linewidth}]{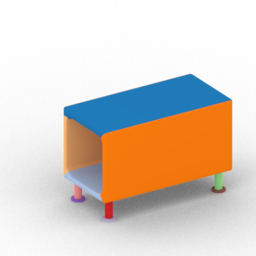} &
        \includegraphics[{width=.14\linewidth}]{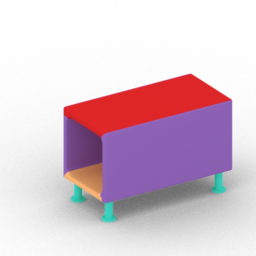} &
        \includegraphics[{width=.14\linewidth}]{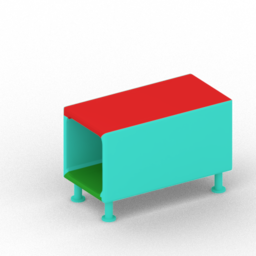} &
        \includegraphics[{width=.14\linewidth}]{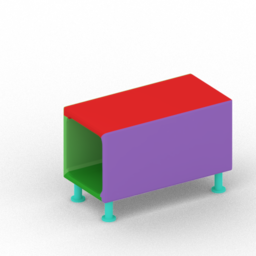} &
        \includegraphics[{width=.14\linewidth}]{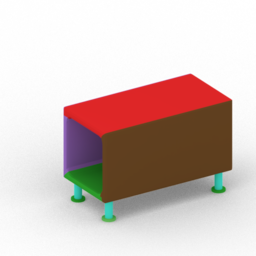} &
        \includegraphics[{width=.14\linewidth}]{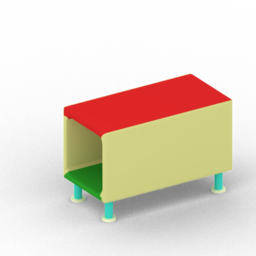} &
        \includegraphics[{width=.14\linewidth}]{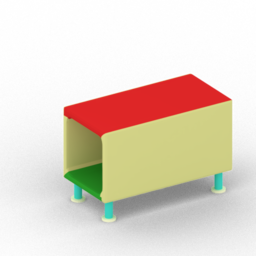} \\
        
        \includegraphics[{width=.14\linewidth}]{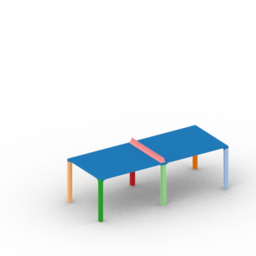} &
        \includegraphics[{width=.14\linewidth}]{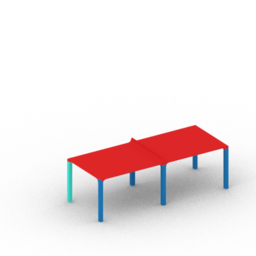} &
        \includegraphics[{width=.14\linewidth}]{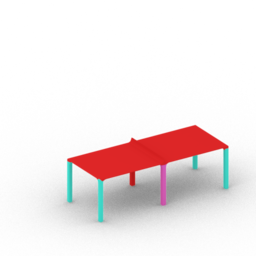} &
        \includegraphics[{width=.14\linewidth}]{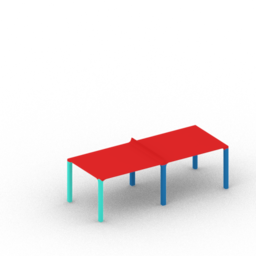} &
        \includegraphics[{width=.14\linewidth}]{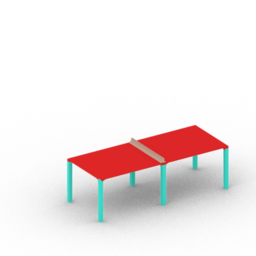} &
        \includegraphics[{width=.14\linewidth}]{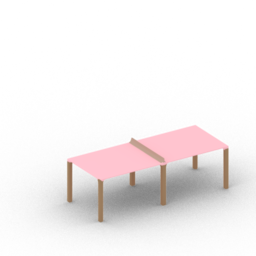} &
        \includegraphics[{width=.14\linewidth}]{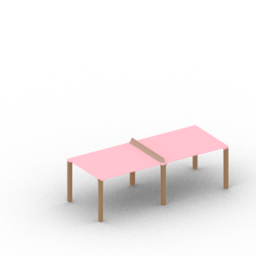} \\
    \end{tabular}
    
    \begin{tabular}{|l|}
\hline
\\
Table $\xrightarrow{}$ game table; picnic table; regular table  \\
Game table $\xrightarrow{}$ ping pong table; pool table  \\
Picnic table $\xrightarrow{}$ bench; bench connector; regular picnic table \\
Regular table $\xrightarrow{}$ regular base; regular tabletop \\
Ping pong table $\xrightarrow{}$ \colorbox{mocha}{net}; ping pong base; ping pong tabletop  \\
Pool table $\xrightarrow{}$ ball; pool base; pool tabletop \\
Regular picnic table $\xrightarrow{}$ picnic base; picnic tabletop \\
Regular base $\xrightarrow{}$ drawer base; pedestal base; regular leg base; star leg base; \\ 
Regular tabletop $\xrightarrow{}$ \colorbox{red_o}{dropleaf}; \colorbox{red_l}{frame}; \colorbox{red_d}{surface} \\
Ping pong base $\xrightarrow{}$ ping pong regular leg base \\
Ping pong tabletop $\xrightarrow{}$ \colorbox{pink}{ping pong surface} \\
Pool base $\xrightarrow{}$ pool regular leg base\\
Pool tabletop $\xrightarrow{}$ frame; surface \\ 
Picnic base $\xrightarrow{}$ picnic regular leg base \\
Picnic tabletop $\xrightarrow{}$ surface \\
Drawer base $\xrightarrow{}$ \colorbox{purple_p}{back panel}; \colorbox{green_l}{bar stretcher}; \colorbox{orange_l}{bottom panel}; \colorbox{gold}{cabinet door}; \colorbox{lime}{caster}; \\  \quad \quad \quad \quad \colorbox{purple_d}{drawer}; \colorbox{mint}{foot}; keyboard tray; \colorbox{green_d}{shelf}; \colorbox{turqoise}{leg}; \colorbox{cardinal}{tabletop connector};  \\
\quad \quad \quad \quad \colorbox{yellow_d}{vertical divider panel}; \colorbox{dark_brown}{vertical front panel}; \colorbox{yellow_l}{vertical side panel} \\
Pedestal base $\xrightarrow{}$ central support; \colorbox{blue_la}{pedestal}; \colorbox{baby_blue}{tabletop connector} \\
Regular leg base $\xrightarrow{}$ \colorbox{teal}{bar stretcher}; caster; \colorbox{iceberg}{circular stretcher}; \colorbox{blue_a}{foot}; \colorbox{blue_n}{leg}; \colorbox{blue_l}{runner}; \colorbox{fush}{tabletop connector} \\
Star leg base $\xrightarrow{}$ central support; star leg set \\
Ping pong regular leg base $\xrightarrow{}$ bar stretcher; \colorbox{cafe}{leg} \\
Picnic regular leg base $\xrightarrow{}$ leg \\
Pool regular leg base $\xrightarrow{}$ leg \\
\\
\hline
\end{tabular}
    
    \caption{Qualitative Results and Grammar for the Table category. }
    \label{fig:table_qual}
\end{figure*}

\begin{figure*}[]
    \centering
    \setlength{\tabcolsep}{1pt}
    \begin{tabular}{ccccccc}
        Input Regions & PartNet (R) & BAE-NET (R) & LEL (R) & LHSS & NGSP & GT
        \\
        \includegraphics[{width=.14\linewidth}]{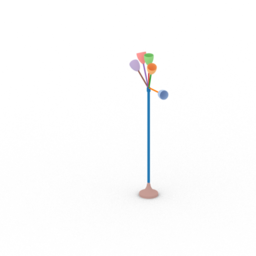} &
        \includegraphics[{width=.14\linewidth}]{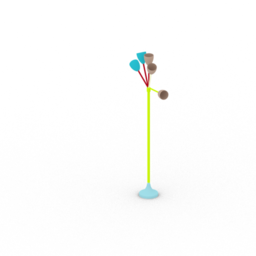} &
        \includegraphics[{width=.14\linewidth}]{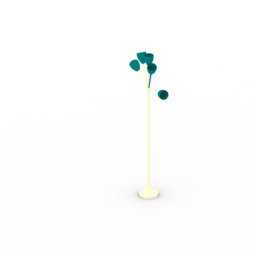} &
        \includegraphics[{width=.14\linewidth}]{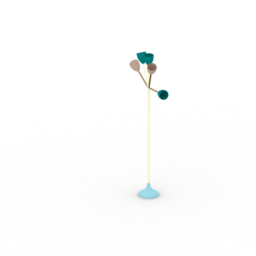} &
        \includegraphics[{width=.14\linewidth}]{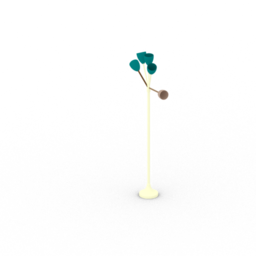} &
        \includegraphics[{width=.14\linewidth}]{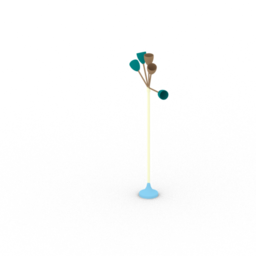} &
        \includegraphics[{width=.14\linewidth}]{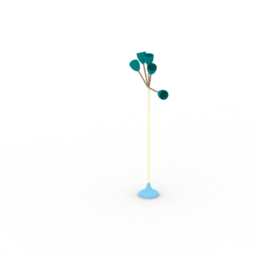} \\
        
        \includegraphics[{width=.14\linewidth}]{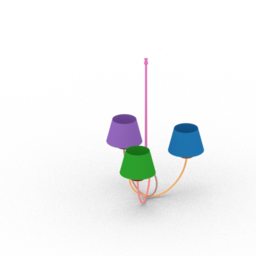} &
        \includegraphics[{width=.14\linewidth}]{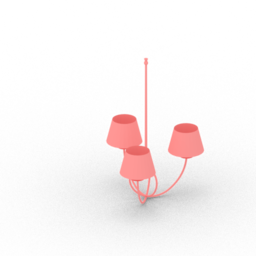} &
        \includegraphics[{width=.14\linewidth}]{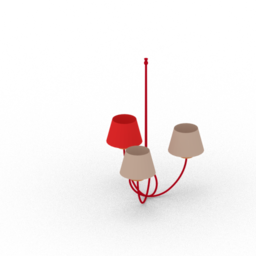} &
        \includegraphics[{width=.14\linewidth}]{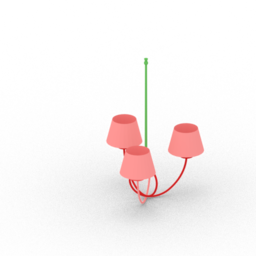} &
        \includegraphics[{width=.14\linewidth}]{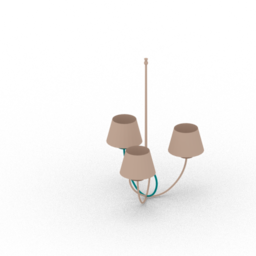} &
        \includegraphics[{width=.14\linewidth}]{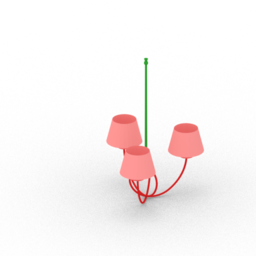} &
        \includegraphics[{width=.14\linewidth}]{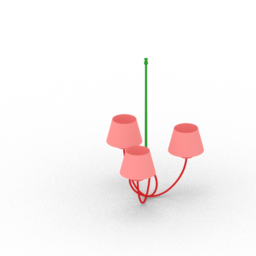} \\
        
        \includegraphics[{width=.14\linewidth}]{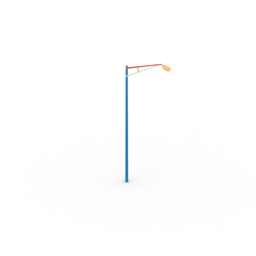} &
        \includegraphics[{width=.14\linewidth}]{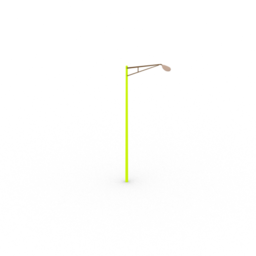} &
        \includegraphics[{width=.14\linewidth}]{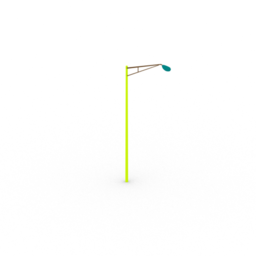} &
        \includegraphics[{width=.14\linewidth}]{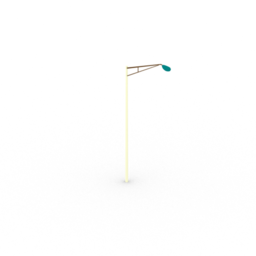} &
        \includegraphics[{width=.14\linewidth}]{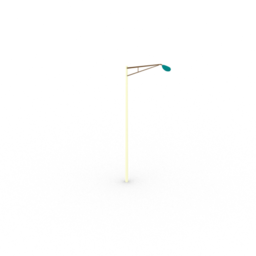} &
        \includegraphics[{width=.14\linewidth}]{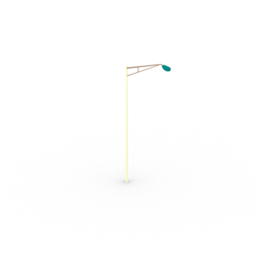} &
        \includegraphics[{width=.14\linewidth}]{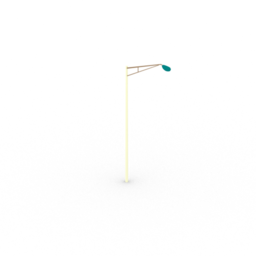} \\
    \end{tabular}
    
    \begin{tabular}{|l|}
\hline
\\
Lamp $\xrightarrow{}$ ceiling lamp; street lamp; table or floor lamp; wall lamp \\
Ceiling lamp $\xrightarrow{}$ chandelier; pendant lamp \\
Street lamp  $\xrightarrow{}$ \colorbox{mint}{post}; street unit; \colorbox{baby_blue}{street base} \\
Table or floor lamp  $\xrightarrow{}$ ToF base; ToF body; ToF unit; ToF power cord \\
Wall lamp  $\xrightarrow{}$ wall base; body; wall unit \\
Chandelier  $\xrightarrow{}$  \colorbox{green_d}{chain}; chandelier base; \colorbox{green_l}{body} ;  chandelier unit group \\
Pendant lamp  $\xrightarrow{}$ pendant base; pendant unit; pendant power cord \\ 
Street unit $\xrightarrow{}$ \colorbox{cafe}{arm}; \colorbox{teal}{head} \\
ToF base $\xrightarrow{}$ ToF holistic base; ToF leg base \\
ToF body $\xrightarrow{}$ \colorbox{yellow_d}{jointed} ; \colorbox{gold}{solid} ; \colorbox{lime}{pole}; vertical panel \\
ToF unit $\xrightarrow{}$ \colorbox{green_d}{connector}; \colorbox{dark_brown}{arm}; \colorbox{mocha}{head} \\
ToF power cord $\xrightarrow{}$ \colorbox{pink}{cord} \\
Wall base $\xrightarrow{}$ wall holistic base \\
Wall unit $\xrightarrow{}$  \colorbox{teal}{arm}; \colorbox{fush}{head} \\
Chandelier base $\xrightarrow{}$ chandelier holistic base \\
Chandelier unit group $\xrightarrow{}$ chandelier unit \\
Pendant base $\xrightarrow{}$ pendant holistic base \\
Pendant unit $\xrightarrow{}$ \colorbox{cardinal}{chain}; \colorbox{orange_l}{head} \\
Pendant power cord $\xrightarrow{}$ cord \\
ToF holistic base $\xrightarrow{}$ \colorbox{blue_l}{base part} \\
ToF leg base $\xrightarrow{}$ \colorbox{blue_n}{leg} \\
Wall holistic base $\xrightarrow{}$ \colorbox{turqoise}{base part} \\
Chandelier holistic base $\xrightarrow{}$ base part \\
Chandelier unit $\xrightarrow{}$ \colorbox{red_d}{arm}; \colorbox{red_l}{head} \\
Pendant holistic base $\xrightarrow{}$ \colorbox{blue_a}{base part} \\
\\
\hline
\end{tabular}
    
    \caption{ Qualitative Results and Grammar for the Lamp category. }
    \label{fig:lamp_qual}
\end{figure*}

\begin{figure*}[]
    \centering
    \setlength{\tabcolsep}{1pt}
    \begin{tabular}{ccccccc}
        Input Regions & PartNet (R) & BAE-NET (R) & LEL (R) & LHSS & NGSP & GT
        \\
        \includegraphics[{width=.14\linewidth}]{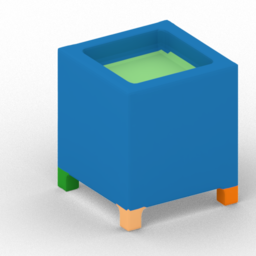} &
        \includegraphics[{width=.14\linewidth}]{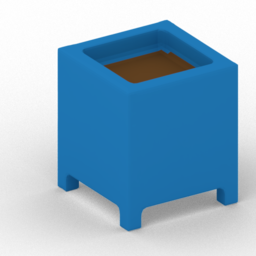} &
        \includegraphics[{width=.14\linewidth}]{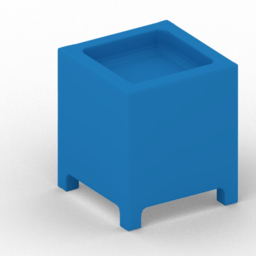} &
        \includegraphics[{width=.14\linewidth}]{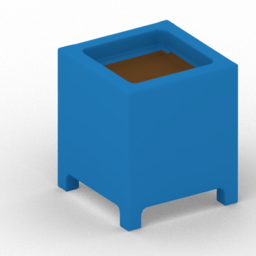} &
        \includegraphics[{width=.14\linewidth}]{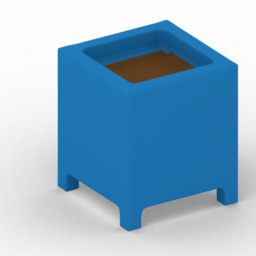} &
        \includegraphics[{width=.14\linewidth}]{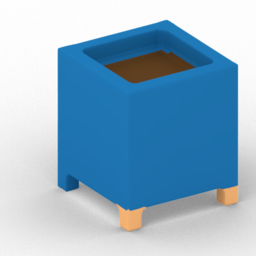} &
        \includegraphics[{width=.14\linewidth}]{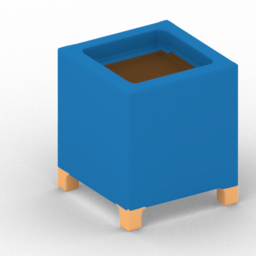} \\
        
        \includegraphics[{width=.14\linewidth}]{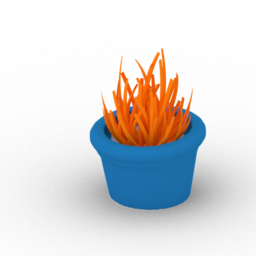} &
        \includegraphics[{width=.14\linewidth}]{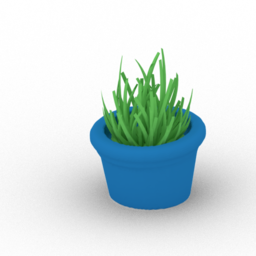} &
        \includegraphics[{width=.14\linewidth}]{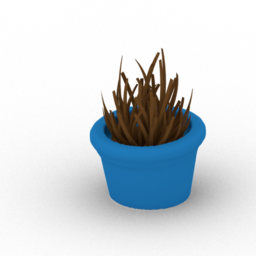} &
        \includegraphics[{width=.14\linewidth}]{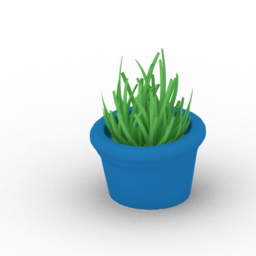} &
        \includegraphics[{width=.14\linewidth}]{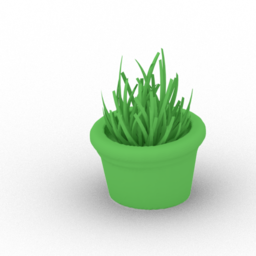} &
        \includegraphics[{width=.14\linewidth}]{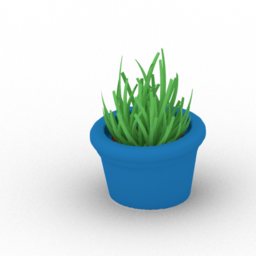} &
        \includegraphics[{width=.14\linewidth}]{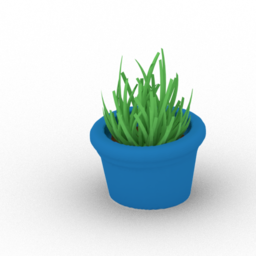} \\
        
        \includegraphics[{width=.14\linewidth}]{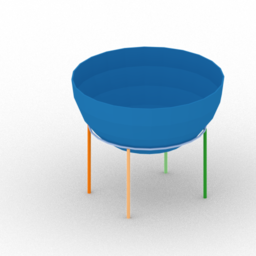} &
        \includegraphics[{width=.14\linewidth}]{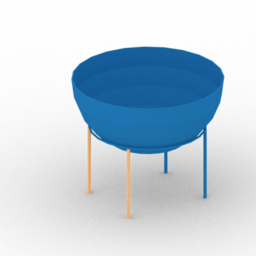} &
        \includegraphics[{width=.14\linewidth}]{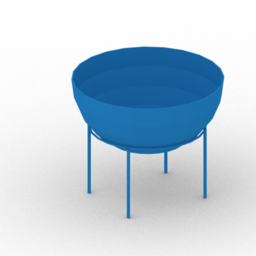} &
        \includegraphics[{width=.14\linewidth}]{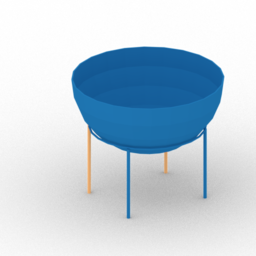} &
        \includegraphics[{width=.14\linewidth}]{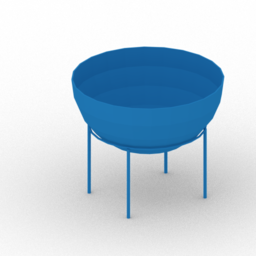} &
        \includegraphics[{width=.14\linewidth}]{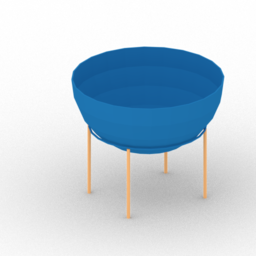} &
        \includegraphics[{width=.14\linewidth}]{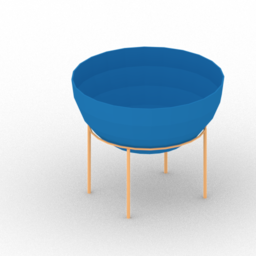} \\
    \end{tabular}
    
    \begin{tabular}{|l|}
\hline
\\
Vase $\xrightarrow{}$ base; body; containing things \\
Base $\xrightarrow{}$ foot base \\
Body $\xrightarrow{}$ \colorbox{blue_n}{container}; lid \\
Containing things $\xrightarrow{}$ \colorbox{dark_brown}{liquid or soil}; \colorbox{green_l}{plant} \\
Foot base \textbf{} \colorbox{orange_l}{foot} \\
\\
\hline
\end{tabular}
    
    \caption{ Qualitative Results and Grammar for the Vase category. }
    \label{fig:vase_qual}
\end{figure*}

\begin{figure*}[]
    \centering
    \setlength{\tabcolsep}{1pt}
    \begin{tabular}{ccccccc}
        Input Regions & PartNet (R) & BAE-NET (R) & LEL (R) & LHSS & NGSP & GT
        \\
        \includegraphics[{width=.14\linewidth}]{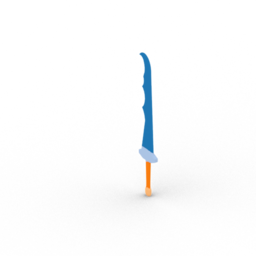} &
        \includegraphics[{width=.14\linewidth}]{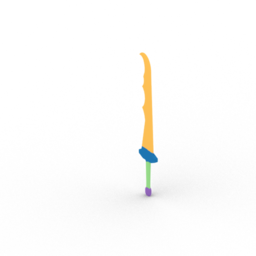} &
        \includegraphics[{width=.14\linewidth}]{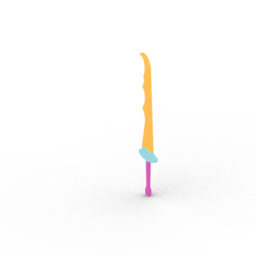} &
        \includegraphics[{width=.14\linewidth}]{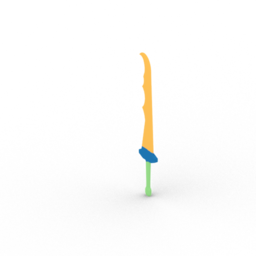} &
        \includegraphics[{width=.14\linewidth}]{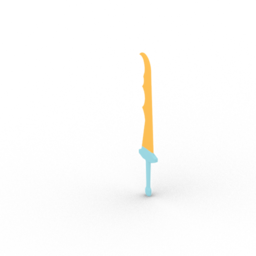} &
        \includegraphics[{width=.14\linewidth}]{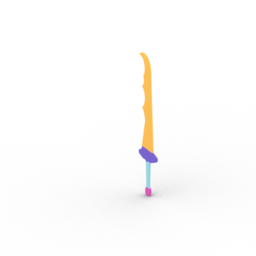} &
        \includegraphics[{width=.14\linewidth}]{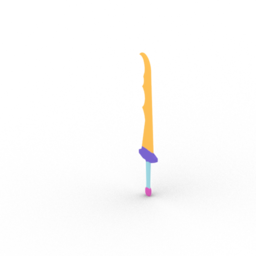} \\
        
        \includegraphics[{width=.14\linewidth}]{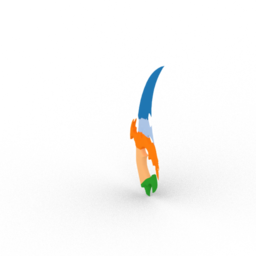} &
        \includegraphics[{width=.14\linewidth}]{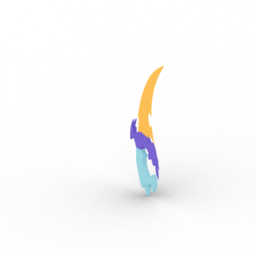} &
        \includegraphics[{width=.14\linewidth}]{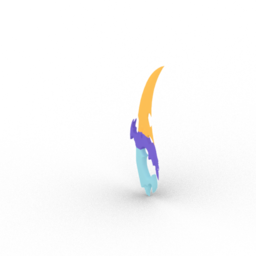} &
        \includegraphics[{width=.14\linewidth}]{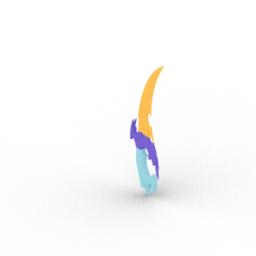} &
        \includegraphics[{width=.14\linewidth}]{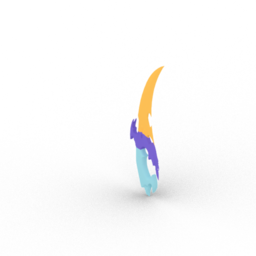} &
        \includegraphics[{width=.14\linewidth}]{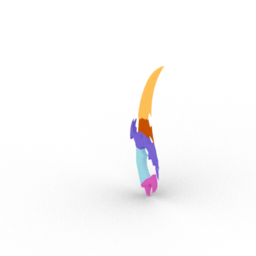} &
        \includegraphics[{width=.14\linewidth}]{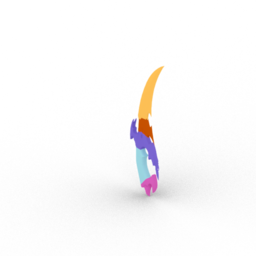} \\
        
        \includegraphics[{width=.14\linewidth}]{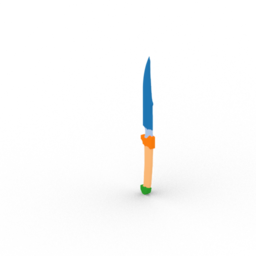} &
        \includegraphics[{width=.14\linewidth}]{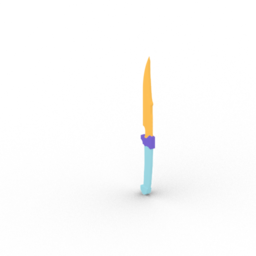} &
        \includegraphics[{width=.14\linewidth}]{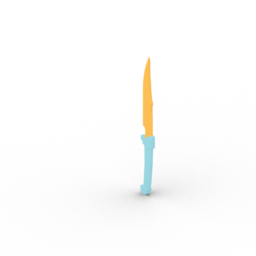} &
        \includegraphics[{width=.14\linewidth}]{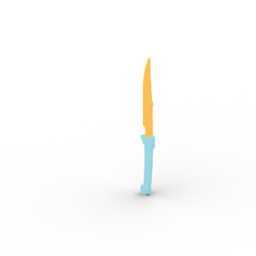} &
        \includegraphics[{width=.14\linewidth}]{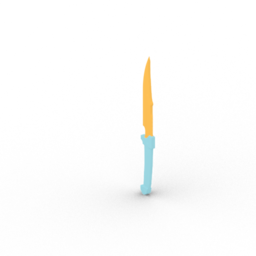} &
        \includegraphics[{width=.14\linewidth}]{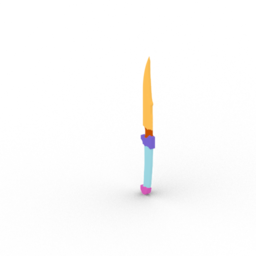} &
        \includegraphics[{width=.14\linewidth}]{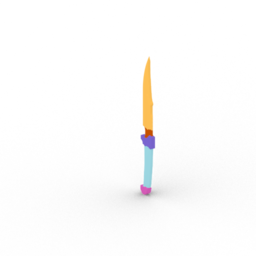} \\
    \end{tabular}
    
    \begin{tabular}{|l|}
\hline
\\
Knife $\xrightarrow{}$ dagger; cutting instrument \\ 
Dagger $\xrightarrow{}$ dagger blade side; dagger handle side \\
Cutting instrument $\xrightarrow{}$ cutting instrument blade side; cutting instrument handle side\\
Dagger blade side $\xrightarrow{}$ \colorbox{orange_l}{blade} \\
Dagger handle side $\xrightarrow{}$ \colorbox{purple_p}{butt}; \colorbox{blue_n}{guard}; \colorbox{green_sl}{handle} \\
Cutting instrument blade side $\xrightarrow{}$ \colorbox{yellow_d}{blade}; \colorbox{red_o}{bolster} \\
Cutting instrument handle side $\xrightarrow{}$ \colorbox{fush}{butt}; \colorbox{purple_d}{guard}; \colorbox{blue_l}{handle} \\
\\
\hline
\end{tabular}

    \caption{ Qualitative Results and Grammar for the Knife category. }
    \label{fig:knife_qual}
\end{figure*}

\begin{figure*}[]
    \centering
    \setlength{\tabcolsep}{1pt}
    \begin{tabular}{ccccccc}
        Input Regions & PartNet (R) & BAE-NET (R) & LEL (R) & LHSS & NGSP & GT
        \\
        \includegraphics[{width=.14\linewidth}]{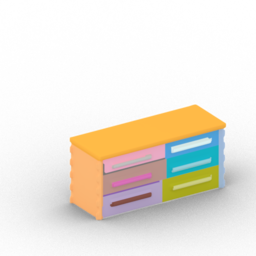} &
        \includegraphics[{width=.14\linewidth}]{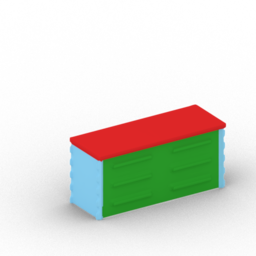} &
        \includegraphics[{width=.14\linewidth}]{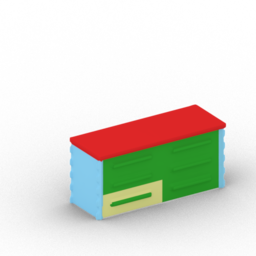} &
        \includegraphics[{width=.14\linewidth}]{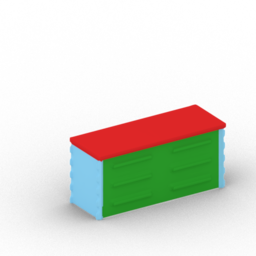} &
        \includegraphics[{width=.14\linewidth}]{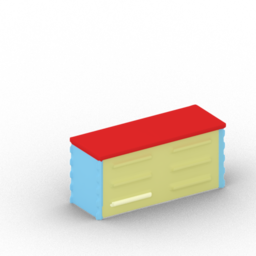} &
        \includegraphics[{width=.14\linewidth}]{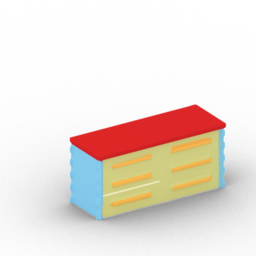} &
        \includegraphics[{width=.14\linewidth}]{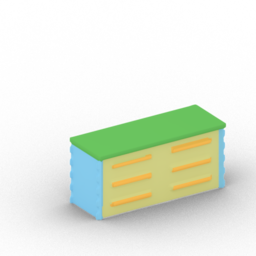} \\
        
        \includegraphics[{width=.14\linewidth}]{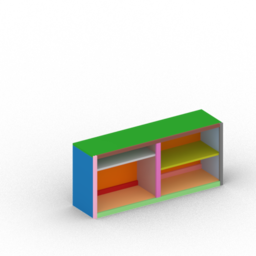} &
        \includegraphics[{width=.14\linewidth}]{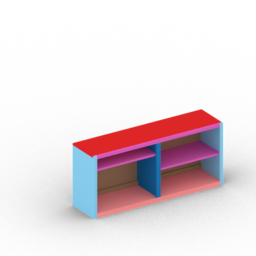} &
        \includegraphics[{width=.14\linewidth}]{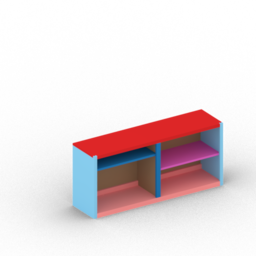} &
        \includegraphics[{width=.14\linewidth}]{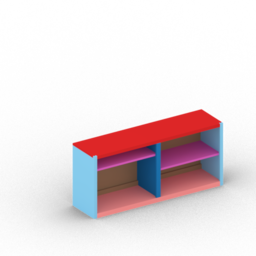} &
        \includegraphics[{width=.14\linewidth}]{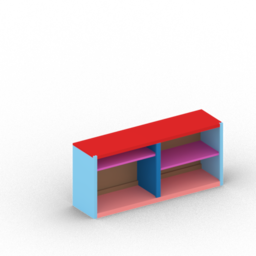} &
        \includegraphics[{width=.14\linewidth}]{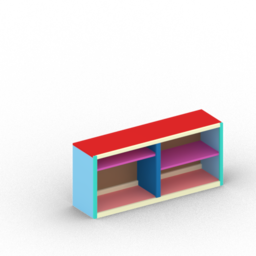} &
        \includegraphics[{width=.14\linewidth}]{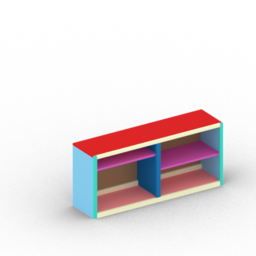} \\
        
        \includegraphics[{width=.14\linewidth}]{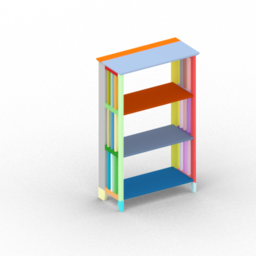} &
        \includegraphics[{width=.14\linewidth}]{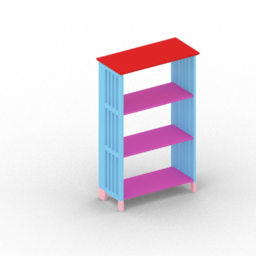} &
        \includegraphics[{width=.14\linewidth}]{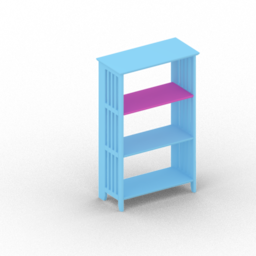} &
        \includegraphics[{width=.14\linewidth}]{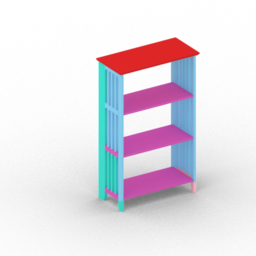} &
        \includegraphics[{width=.14\linewidth}]{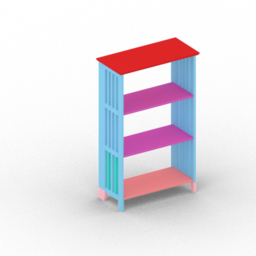} &
        \includegraphics[{width=.14\linewidth}]{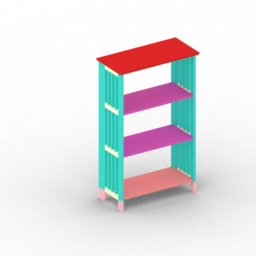} &
        \includegraphics[{width=.14\linewidth}]{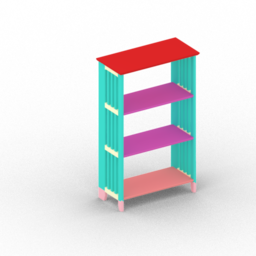} \\
    \end{tabular}
    
    \begin{tabular}{|l|}
\hline
\\ 
Storage Furniture $\xrightarrow{}$ base; \colorbox{green_d}{door}; frame; \colorbox{green_l}{countertop}; drawer; \colorbox{fush}{shelf} \\
Base $\xrightarrow{}$ \colorbox{pink}{foot base}; panel base \\
Frame $\xrightarrow{}$ \colorbox{cafe}{back panel}; \colorbox{red_l}{bottom panel}; \colorbox{mint}{horizontal bar}; \colorbox{turqoise}{vertical bar}; \colorbox{red_d}{top panel}; \\
\quad \quad \quad \quad \colorbox{blue_n}{vertical divider panel}; \colorbox{blue_l}{vertical front panel}; \colorbox{baby_blue}{vertical side panel} \\
Drawer $\xrightarrow{}$ drawer box; \colorbox{yellow_d}{handle} \\
Drawer Box $\xrightarrow{}$ back; bottom; \colorbox{yellow_l}{front}; side \\
\\
\hline
\end{tabular}
    
    \caption{ Qualitative Results and Grammar for the Storage Furniture category. }
    \label{fig:storage_qual}
\end{figure*}

We present additional qualitative results comparing different methods on the task of fine-grained semantic segmentation of Partnet shapes for Chairs (Figure \ref{fig:chair_qual}), Tables (Figure \ref{fig:table_qual}), Lamps (Figure \ref{fig:lamp_qual}), Vases (Figure \ref{fig:vase_qual}), Knives (Figure \ref{fig:knife_qual}) and Storage Furniture (Figure \ref{fig:storage_qual}).

Each figure additionally contains the semantic grammar we use.
All shape grammars we use are derived from the hierarchies defined by PartNet. In most cases these labels corresponds to the \textit{level-2} granularity in PartNet. For some shapes, where level-2 was not defined, level-3 was substituted. 
For all terminal labels that are present in the depicted qualitative examples, we color the background text of the terminal label to match the color of semantic part in the qualitative renders.
The input region column is given purely random colors, such that each shape region is given a unique color.

\end{document}